\documentclass[10pt,twocolumn,letterpaper]{article}

 \usepackage[pagenumbers]{cvpr} 

\usepackage{multirow}  		
\usepackage{amsmath} 		
\usepackage{wrapfig} 		
\usepackage{graphicx} 		
\usepackage{subcaption} 	
\usepackage{bm}
\usepackage{algorithm}
\usepackage[noend]{algorithmic} %
\usepackage{colortbl}
\definecolor{cvprblue}{rgb}{0.21,0.49,0.74}
\usepackage[pagebackref,breaklinks,colorlinks,allcolors=cvprblue]{hyperref}

\def\confName{CVPR}
\def\confYear{2025}

\title{Learning Class Prototypes for Unified Sparse Supervised 3D Object Detection}

\author{Yun Zhu\textsuperscript{\rm 1}, ~Le Hui\textsuperscript{\rm 2},~Hang Yang\textsuperscript{\rm 1}, ~Jianjun Qian\textsuperscript{\rm 1}, Jin Xie\textsuperscript{\rm 3,4}\thanks{Corresponding author}, ~Jian Yang\textsuperscript{\rm 1}\\
	\textsuperscript{\rm 1}PCA Lab, Nanjing University of Science and Technology, Nanjing, China\\ 
	\textsuperscript{\rm 2}School of Electronics and Information, Northwestern Polytechnical University, Xi’an, China \\
	\textsuperscript{\rm 3}State Key Laboratory for Novel Software Technology, Nanjing University, Nanjing, China \\
	\textsuperscript{\rm 4}School of Intelligence Science and Technology, Nanjing University, Suzhou, China \\
	\tt\small \{zhu.yun, hangyang, csjqian, csjyang\}@njust.edu.cn; csjxie@nju.edu.cn; huile@nwpu.edu.cn
}

\begin{document}
\maketitle
\begin{abstract}
Both indoor and outdoor scene perceptions are essential for embodied intelligence. However, current sparse supervised 3D object detection methods focus solely on outdoor scenes without considering indoor settings.
To this end, we propose a unified sparse supervised 3D object detection method for both indoor and outdoor scenes through learning class prototypes to effectively utilize unlabeled objects.
Specifically, we first propose a prototype-based object mining module that converts the  unlabeled object mining into a matching problem between class prototypes and unlabeled features. By using optimal transport matching results, we assign prototype labels to high-confidence features, thereby achieving the mining of unlabeled objects.
We then present a multi-label cooperative refinement module to effectively recover missed detections through pseudo label quality control and prototype label cooperation.
Experiments show that our method achieves state-of-the-art performance under the one object per scene sparse supervised setting across indoor and outdoor datasets. 
With only one labeled object per scene, our method achieves about 78\%,  90\%, and  96\% performance compared to the fully supervised detector on ScanNet V2, SUN RGB-D, and KITTI, respectively, highlighting the scalability of our method. Code is available at
\emph{\url{https://github.com/zyrant/CPDet3D}}.
\end{abstract}
    
\definecolor{light green}{RGB}{0,224,0}
\definecolor{light orange}{RGB}{255,175,0}

\section{Introduction}
\label{sec:intro}

3D object detection attracts more and more attention, as it is crucial for the vision tasks such as autonomous driving and emboddied robotics~\cite{wu2024text2lidar, hui20223d, zhang20243d}. Conventional 3D object detection~\cite{votenet, cagroup, voxelnet, second, pillarnet} relies on a large number of precise annotations,  which are labor-intensive and time-consuming. How to achieve efficient and accurate 3D object detection under limited annotations has become an important research topic in the field of 3D object detection.

\begin{figure}
	\centering
	\includegraphics[width=1.0\linewidth]{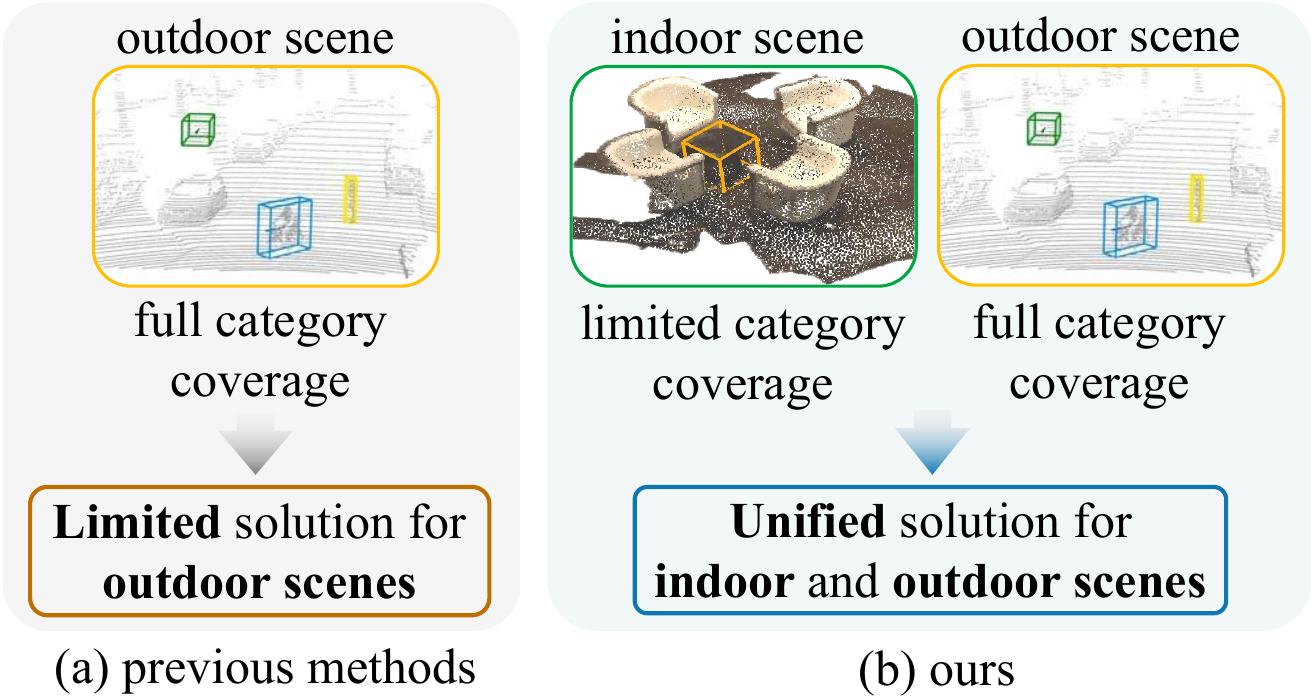}
	\caption{Comparison of sparse supervised 3D object detection methods. 
	 Previous methods rely on the premise of full category coverage within each scene, achieved through a GT sampling strategy, as represented in (a). This approach encounters limitations in indoor scenes, which have scene-specific categories. In contrast, we propose a unified sparse 3D object detection scheme (b) applicable to both indoor and outdoor scenes, utilizing nearest prototype retrieval for effective object mining.}
	\label{fig:unvier}
	\vspace{-15pt}
\end{figure}

Current 3D object detection methods designed for limited annotations can be roughly categorized into weakly supervised~\cite{weakly3d,backtoreal}, semi supervised~\cite{dqs3d,dual,ioumatch,sess}, and sparse supervised methods~\cite{ss3d, coin}. Weakly supervised methods use weaker supervision than box annotations as constraints, such as point-level annotations (\textit{i.e.}, box centers). However, point-level annotations cannot provide precise box attributes. These works require a small number of precise annotations~\cite{weakly3d} or synthetic 3D shapes~\cite{backtoreal}. Semi supervised 3D object detection naturally involves precise annotations in some scenes of the datasets, while the rest remain unlabeled. These types of works achieve better performance than point supervision. Nonetheless, semi supervision faces two drawbacks. First, there is a domain gap between labeled and unlabeled scenes, which makes information transfer ineffective when the gap is large. Second, annotating an entire scene is hindered by its complexity and density, making the process labor-intensive.

\begin{figure}
	\centering
	\includegraphics[width=1.0\linewidth]{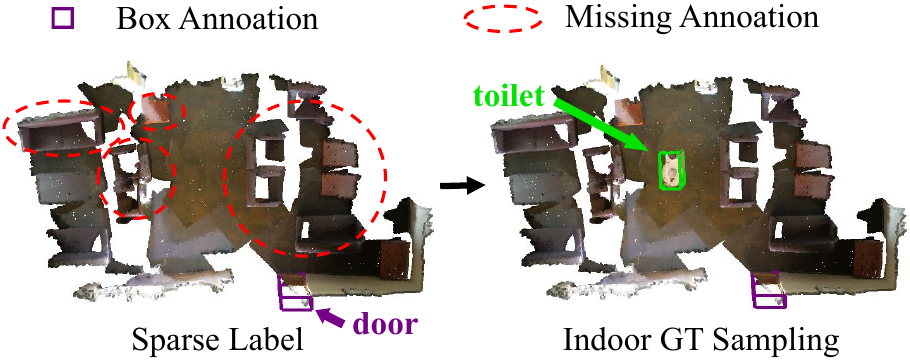}
	\caption{Visualization of GT sampling on indoor dataset ScanNet V2. Unlike outdoor autonomous driving scenes, indoor objects have specific classes, making the use of GT sampling impractical. As shown in the image, placing a bathroom-specific ``toilet" in the living room is unreasonable.}
	\label{fig:indoor_copy}
	\vspace{-15pt}
\end{figure}

To reduce domain gap and labor cost, sparse supervised 3D object detection annotates a limited number of annotations within each scene. However, sparse supervised 3D object detection still faces the challenge of a large number of missing annotations within scenes. Current sparse supervised 3D object detection methods~\cite{ss3d, coin} are specifically designed for outdoor scenes and use a GT Sampling strategy~\cite{second} to ensure all categories (\textit{i.e.}, car, pedestrian, and bicycle) are present in a single scene, as shown in  Fig.~\ref{fig:unvier}. With such a strategy, they can mine the unlabeled objects of all categories from an individual scene. In contrast,  indoor scenes contain scene-specific categories and it would be unreasonable to place a bathroom specific ``toilet" in the living room, as shown in Fig.~\ref{fig:indoor_copy}. Therefore, existing methods are primarily applied to outdoor scenarios and face limitations when handling indoor environments, which restricts their applicability.

In this paper, we propose a sparse supervised 3D object detection method that can achieve impressive results in both indoor and outdoor environments through learning class prototypes to effectively leverage the unlabeled objects.
Specifically, the method contains two modules: a prototype-based object mining module and a multi-label cooperative refinement module.
The first module formulates the object mining problem as an optimal transport matching task between clustered class prototypes and unlabeled features. Initially, we generate class-aware prototypes that represent the class aware feature representations by clustering the features of labeled objects across scenes. Then, by applying optimal transport for matching, we establish a correspondence between these prototypes and unlabeled features beyond a single scene. Based on the matching results, we assign category labels (prototype labels) to high-confidence features, enabling the mining of unlabeled objects. 
The second module includes iterative pseudo labeling and prototype label cooperating. The iterative pseudo labeling generates high-quality pseudo labels by filtering out inaccurate predictions, while the prototype label cooperating builds on the above process to fill in missed detections by assigning prototype labels to undetected objects.
Our contributions are listed as follows:

\begin{itemize}

\item We propose a unified solution for sparse supervised 3D object detection, effective across both indoor and outdoor environments.

\item We develop a prototype-based object mining module that assigns prototype labels to unlabeled objects beyond scene limitations and a  multi-label refinement module that recovers missed detections with pseudo and prototype labels in a self-training approach.

\item Extensive results show that our method achieves 78\% and 90\% performance compared to the fully supervised detector with only 7\% and 26\% annotations on ScanNet V2~\cite{scannet} and SUN RGB-D~\cite{sunrgbd}. Besides, experiments on outdoor KITTI show that our method achieves up to 96\% of fully supervised performance under sparse supervision.

\end{itemize}
 \section{Related Work}
\label{sec:relatedwork}

~~~~~\textbf{3D Object Detection with Full Supervision.} 
For point-based methods, VoteNet \cite{votenet} adopts deep hough voting to group features and generate proposals. As a follow-up work, 
RBGNet \cite{rgbnet} introduces a ray-based feature grouping module to capture points on the object surface features. For voxel-based methods~\cite{fcaf3d, cagroup, tr3d, spgroup3d, pointpillars, second, pillarnet, centerpoint, voxelrcnn}, CenterPoint~\cite{centerpoint} proposes a center-based head for single-stage detection. PillarNet~\cite{pillarnet} proposes an efficient method for gathering features on pillars. FCAF3D \cite{fcaf3d} introduces the anchor-free style into 3D indoor object detection. Based on this, TR3D \cite{tr3d} further improves the results of with a simpler structure. 
SPGroup3D~\cite{spgroup3d} introduces the superpoints to express instance consistency.
Although these methods have achieved high performance, they rely on intensive labeling, highlighting the need for more efficient alternatives to reduce manual annotations.
\begin{figure*}[htbp]
	\centering
	\includegraphics[width=1.0\textwidth]{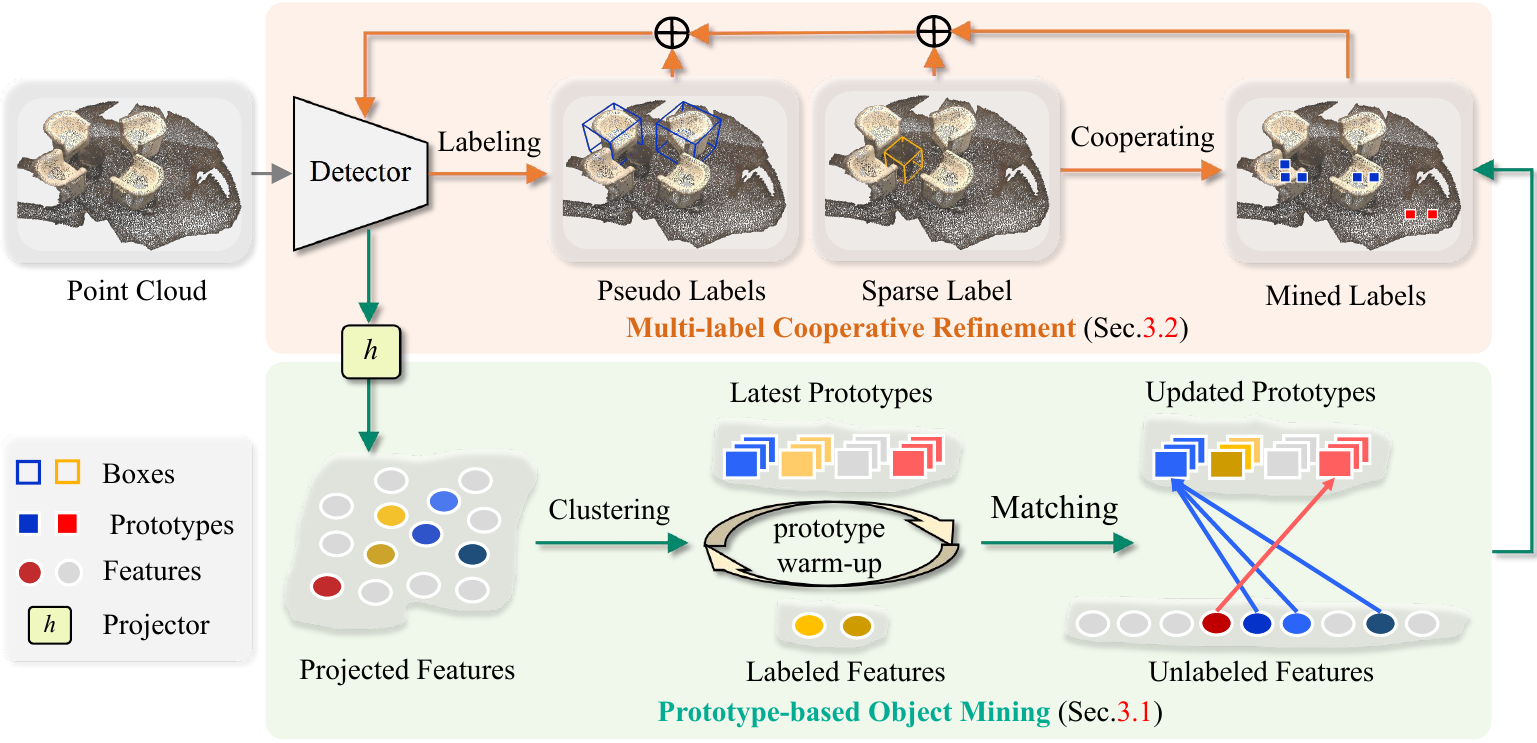}
	\caption{The architecture of our method for sparse supervised 3D object detection is as follows. Given a point cloud and a detector, we first project the features from the detector and cluster them into class-aware prototypes. Based on the learned similarity between prototypes and features, we assign pseudo labels to unlabeled objects. Next, we introduce an effective refinement module that cooperatively utilizes sparse, pseudo, and prototype labels to reduce missed detections during iterative training.}
	\label{fig:framework}
	\vspace{-10pt}
\end{figure*}

\textbf{3D Object Detection with Limited Supervision.} 
Similar to the 2D counterpart~\cite{sparsedet, comining}, limited supervision in 3D area~\cite{tang2022learning} is gaining growing attention. 
Weakly3D~\cite{weakly3d} uses partially precise box annotations and BacktoReal~\cite{backtoreal} adopts synthetic datasets.
SESS~\cite{sess} introduces the first semi supervised 3D object detection using the mean-teacher framework~\cite{meanteacher}. 3DIoUMatch~\cite{ioumatch} adds IoU estimation for pseudo-label filtering, while DQS3D~\cite{dqs3d} adopts a dense match strategy~\cite{denselearn, denseteacher, densepse}. DPKE~\cite{dual} proposes the dual-perspective knowledge enrichment.
For sparse supervised settings, SS3D~\cite{ss3d} uses GT Sampling to increase object count in sparse scenes, while Xia et al.~\cite{coin} leveraged BEV features for object mining~\cite{siod} and further improved its performance with a mix-feature enhancement~\cite{hinted}.
However, outdoor GT Sampling does not suit indoor scenes, as objects like ``toilet" can’t be copied into living room. In this paper, we introduce the first attempt to addresses sparse supervised 3D object detection across indoor and outdoor environments.

\textbf{Prototype-based Learning.} 
Cluster based methods~\cite{hui2021superpoint, hui2022learning}, particularly in 2D semantic segmentation~\cite{visual, exploring, uncovering, rethinkseg, lv2024noise, lv2024context, liu2025milnet}, prototypes help to cluster distinctive features. ContrastiveSeg~\cite{exploring} enhances global pixel relationships, while ProtoSeg~\cite{rethinkseg} optimizes pixel-prototype correspondence. 
In 3D detection, Prototypical VoteNet~\cite{prototypical} applies prototypes for few-shot learning. Unlike the methods mentioned above that use prototypes during both training and testing, our method uses prototypes only during training, thereby  not impacting the testing phase.

\section{Method}
\label{sec:method}
In this section, we introduce our method in three parts. First, in Sec.~\ref{sec:pimine}, we introduce the prototype-based object mining module, which is designed to identify unlabeled objects across diverse scenes using class-aware prototypes. Then, in Sec.~\ref{sec:mcrefine}, we describe the multi-label cooperative refinement module, which integrates pseudo labels with prototype labels to reduce missed detections and improve detection accuracy. Finally, we describe the training strategy for our framework in Sec.~\ref{sec:lo}. An overview of our method is shown in Fig. \ref{fig:framework}.

\subsection{Prototype-based Object Mining}\label{sec:pimine}

Due to the rich geometric information in point clouds, only a small amount of precise bounding boxes are required to achieve satisfactory performance compared to category labels~\cite{mixsup}. In the sparse 3D object detection, some precise bounding box annotations are already provided. Therefore, we focus on mining category labels to further improve detection performance. 
In indoor scenes, it is difficult to acquire the features of all categories using a single scene under one object per scene sparse supervised setting.
To address this, we propose a prototype-based object mining module to identify unlabeled objects. This includes class-aware prototype clustering to learn class-specific prototypes across scenes and prototype label matching to assign labels based on these prototypes.

\textbf{Class-aware Prototype Clustering.} \label{sec:cpcluster} In order to acquire the features of all categories in datasets, we introduce the class-aware prototype clustering to explore cross-scene semantic feature distributions. 

\begin{algorithm}[t]
	\caption{Class-aware Prototype Clustering.}
	\label{algo:cpcluster}
	\begin{algorithmic}[1]
		\STATE \textbf{Input:} Proposal features $\bm{X} \in \mathbb{R}^{N \times C}$, initial class-aware prototypes $\bm{P} \in \mathbb{R}^{K \times O \times C}$;
		\STATE \textbf{Output:} Updated prototypes $\bm{P'} \in \mathbb{R}^{K \times O \times C}$.
		\STATE { \color{gray}// Projector with MLPs }
		\STATE  $\bm{F} = \text{Projector}(\bm{X})$
		\STATE { \color{gray} // Class-aware cluster and prototype update}
		\FOR{$k\leftarrow 0$ to $K$}
		\IF {$k$ is valid class ($M_k$!= 0)}
		\STATE {\color{gray} // Obtain class features and prototypes}
		\STATE  $\bm{F}_k =\bm{F} * \bm{M}_k$,  Index $\bm{P}_k$
		\STATE {\color{gray}// Compute mapping by Sinkhorn-Knopp}
		\STATE  ${\bm{L}}_k  = \texttt{diag}(\bm{u}) \exp \big(\frac{\bm{P}_k^{\top\!}\bm{F}_k}{\kappa}\big)\texttt{diag}(\bm{v})$
		\FOR {$i\leftarrow 0$ to $O$}
		\STATE {\color{gray}// Update the prototypes with $\mu$}
		\STATE  $\bm{p'}_{k,i} \leftarrow \mu\bm{p}_{k,i} + (1-\mu) \frac{1}{N_k} {\sum_{i=1}^{N_k}}\bm{F}_{k,i}$
		\ENDFOR
		\ELSE
		\STATE	Continue
		\ENDIF
		\ENDFOR
		\STATE \textbf{Return $\bm{P'}$}
	\end{algorithmic}
\end{algorithm}

Given a point cloud,  it is fed into the detector to obtain proposal features. Define the proposal features from the detector as $\bm{X} \in \mathbb{R}^{N \times C}$, where \textit{N} and \textit{C} represent the number of proposals and the feature dimension, respectively. We first use a projector consisting of several multi-layer perceptrons~(MLPs) to obtain the projected features $\bm{F} \in \mathbb{R}^{N \times C}$ and enhance the feature diversity. 
After that, based on the projected features $\bm{F}$, we leverage the labeled features to build class-aware prototypes. Suppose there are $K$ categories and take the $k$-th category as an example, we obtain the semantic features $\bm{F}_k \in \mathbb{R}^{M \times C}$ ($M$ $<<$ $N$) corresponding to $k$-th category through the class-aware mask $\bm{M}_k \in \mathbb{R}^{N \times 1}$. This process can be formulated as follows:
\begin{equation}
	\begin{aligned}
		\bm{F}_k =\bm{F} * \bm{M}_k
	\end{aligned}
\end{equation}

$\bm{M}_k$ is a boolean mask to select the true positive projected features. 
Suppose the prototypes of the $k$-th category are $\bm{P}_k \in \mathbb{R}^{O \times C}$, where $O$ $>$ 1 ensures the feature diversity of the same category. Next, we need to obtain the matching matrix between $M$ semantic features and $O$ prototypes so that we can update the prototypes. We model the matching between prototypes and features as an optimal transport problem and choose the Sinkhorn-Knopp iteration~\cite{sinkhorn, exploring} to obtain the feature-to-prototype matching matrix $\bm{L}_k \in \mathbb{R}^{M \times O}$. The solver can be given as follows:

\begin{equation}
	\begin{aligned}
		{\bm{L}}_k  = \texttt{diag}(\bm{u}) \exp \big(\frac{\bm{P}_k^{\top\!}\bm{F}_k}{\kappa}\big)\texttt{diag}(\bm{v})
	\end{aligned}
\end{equation}
where $\bm{u} \in \mathbb{R}^{M}$ and $\bm{v} \in \mathbb{R}^{O}$ are renormalization vectors, computed by a few steps of Sinkhorn-Knopp iteration. ${\bm{P}_k^{\top\!}\bm{F}_k} \in [0,1]$ represents the cosine similarity between prototypes and semantic features.
Based on the matching matrix $\bm{L}_k \in \mathbb{R}^{M \times O}$, we update each prototype with momentum toward the mean of the clustered embeddings. Hence, the $i$-th prototype of class $k$ is updated as:
\begin{equation}
	\begin{aligned}
		\bm{p'}_{k,i} \leftarrow \mu\bm{p}_{k,i} + (1-\mu) \frac{1}{N_k} {\sum_{i=1}^{N_k}}\bm{F}_{k,i}
	\end{aligned}
\end{equation}
where $\bm{F}_{k,i}$ represents one of the $l_2$-normalized projected features being mapped to the $i$-th prototype of class $k$. $\mu \in [0,1]$ is a momentum coefficient. In each iteration, the updated $p'$ is assigned as the new $p$ for the following iteration. Finally, we introduce a prototype-feature contrastive loss~\cite{momentum} as the specific objective function for this module. This loss forces features belonging to the same prototypes to be closer while pushing features with different prototypes away. 
Algorithm \ref{algo:cpcluster} summarizes our class-aware prototype clustering.

\begin{figure}[t]
	\centering
	\includegraphics[width=1.0\linewidth]{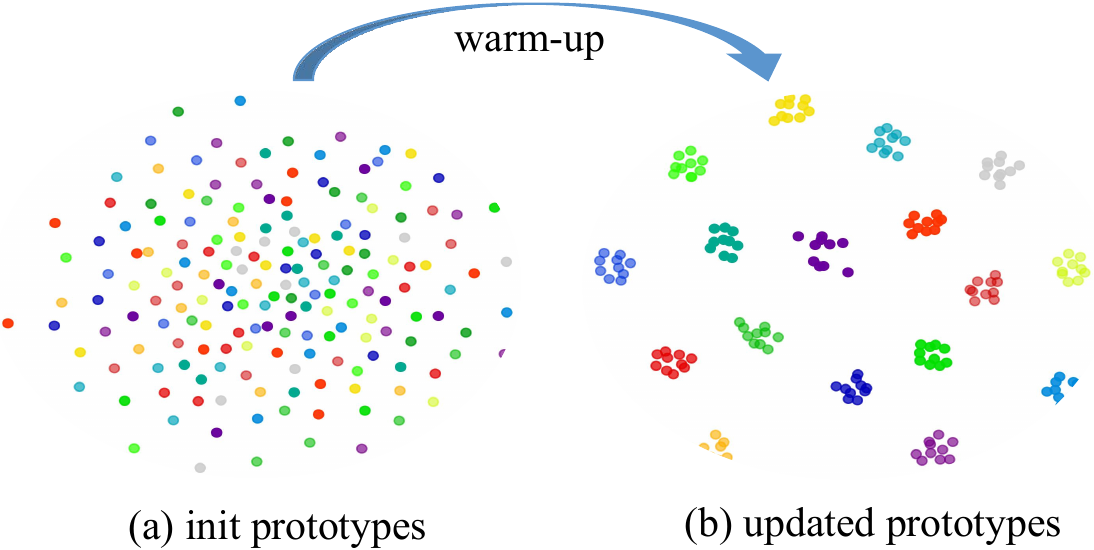}
	\caption{t-SNE results of class-aware prototypes before and after warm-up on ScanNet V2.}
	\label{fig:prototype}
	\vspace{-10pt}
\end{figure}

\textbf{Prototype Label Matching.}\label{sec:plmatch} After obtaining the class-aware prototypes, we propose a prototype label object matching module for unlabeled objects label matching. However, as shown in Fig.~\ref{fig:prototype}, the initial prototypes are initialized using a truncated normal distribution, making them indistinguishable across classes. If we use such prototypes for label matching, this will lead to the confusion of network learning.
Therefore, we set warm-up iterations before label matching.

\begin{figure}
	\centering
	\includegraphics[width=1.0\linewidth]{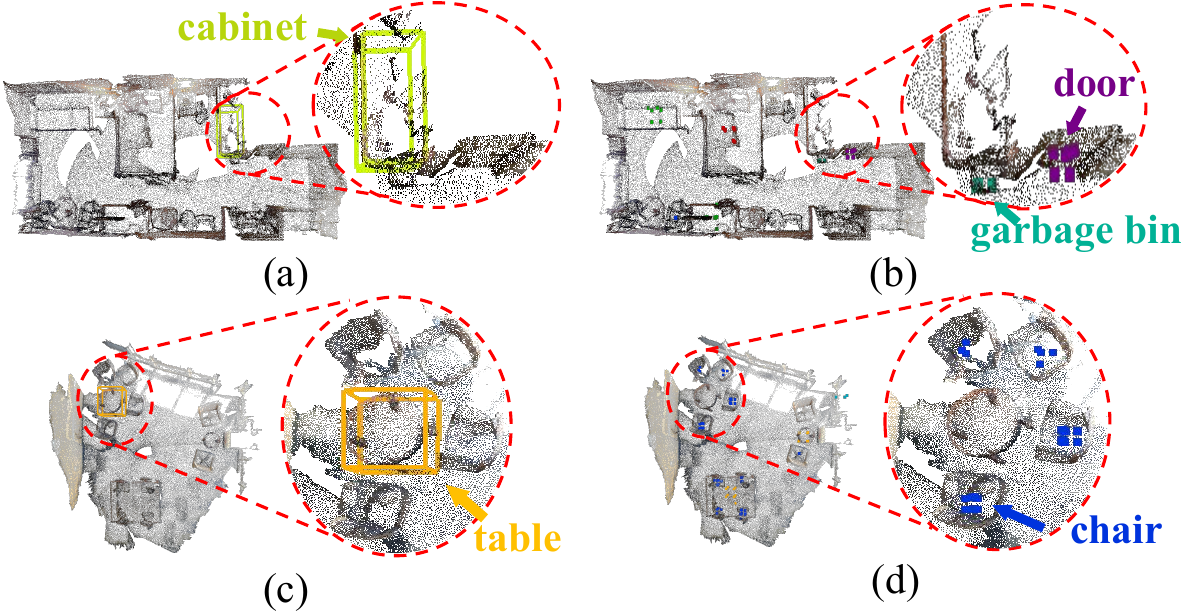}
	\caption{Visualization of real mined prototype labels. The colors in detailed views denote the object categories including chair, table, door, garbage bin, and cabinet. To highlight the labels, we sparsify the point clouds and visualize different types of labels separately. The first column (a) and (c) represent the labeled one object per scene. The second column (b) and (d) represent the real mined prototype labels during training.}
	\label{fig:pomine}
	\vspace{-12pt}
\end{figure}

After the warm-up, we obtain the distinguishable class-aware prototypes $\bm{P} \in \mathbb{R}^{K \times O \times C}$ and projected features $\bm{F} \in \mathbb{R}^{N \times C}$ of the $j$-th scene. 
For the $j$-th scene, we attain an affinity matrix $\bm{A}={\bm{F}^{\top\!}\bm{P} \in \mathbb{R}^{N \times K \times O}}$ of projected features and class-aware prototypes via a dot-product operation and assume that its classification scores from the detector are $\bm{S} \in \mathbb{R}^{N \times K}$. 
To realize label propagation for unlabeled objects, we use the $\bm{S}$ and  $\bm{A'} \in \mathbb{R}^{N \times K}$ to compute the propagation probability $\bm{W} \in \mathbb{R}^{N \times K}$ for the $j$-th scene, which is formulated as:
\begin{equation}
	\begin{aligned}
		\bm{W} &= \bm{S} \odot \bm{A'}, \\
		\texttt{where}~ \bm{A'} &= \mathop{\operatorname{argmax}}\limits_{i=1,\ldots,O}(\bm{A})
	\end{aligned}
\end{equation}
where $\odot$ is the Hadamard product and $\bm{W}$ indicates the probability of features belonging to a certain kind of category. $\bm{A'}$ represents the affinity matrix obtained by assigning each feature to one of the prototypes in each category based on the highest probability. In this module, we consider both the semantic information and feature similarities simultaneously. Based on $\bm{W}$, we derive the category labels $\bm{C}_{f} \in \mathbb{R}^{N \times 1}$ for features by:
\begin{equation}
	\begin{aligned}\label{eq:fakeclass}
		\bm{C}_{f}=\mathop{\operatorname{argmax}}\limits_{i=1,\ldots,K}(\bm{W})
	\end{aligned}
\end{equation}
In Eq. (\ref{eq:fakeclass}), each feature in the $j$-th scene is assigned a category based on the highest probability in each row of $\bm{W}$. Because these category labels are generated by the class-aware prototypes, we call them prototype labels for short to distinguish the categories of the true sparse labels.
After we obtain the $\bm{C}_{f}$, it is essential to filter out three special regions including background regions, regions with true sparse labels, and regions outside of the input point cloud range, as they may affect the detection accuracy. Specifically, we derive the foreground boolean mask $\bm{M}_f \in \mathbb{R}^{N \times 1}$ using a score threshold $\alpha_{pro}$ based on classification score, the true label boolean mask $\bm{M}_s \in \mathbb{R}^{N \times 1}$ based on sparse labels, and the range boolean mask $\bm{M}_r \in \mathbb{R}^{N \times 1}$ computed based on the specified point cloud input range. Therefore, the remaining prototype labels are obtained by:
\begin{equation}
	\begin{aligned}
		\bm{C}{'_p} =\bm{C}_{p} * \bm{M}_f * \bm{M}_s * \bm{M}_r
	\end{aligned}
\end{equation}
where $\bm{C}{'_p \in \mathbb{R}^{E \times 1}}$ ($E$ $<<$ $N$) denotes the remaining prototype labels and we set a prototype classification loss in this module to take advantage of them. The real mined prototype labels during training are shown in Fig. \ref{fig:pomine}. It is evident that our module can identify the unlabeled objects in the scenes, even if these categories are not labeled in the scenes.

\subsection{Multi-label Cooperative Refinement}\label{sec:mcrefine}
Iterative training improves the performance in limited annotation settings by relying on pseudo labels. Generally speaking, high thresholds ensure the quality of pseudo labels, but also lead to missed detections. Existing methods solve this problem by dynamically adjusting the threshold in complex ways~\cite{calibrated} to achieve a balance. Here, we propose an  effective solution to address this problem by integrating sparse, pseudo, and prototype labels.
 
\textbf{Iterative Pseudo Labeling.} The first thing is to generate high-quality pseudo labels. This process involves the following steps. 
Suppose the one stage detector generates the predictions $y_j$ of the $j$-th input point cloud scene. We establish a classification score threshold, $\alpha_{cls}$, to eliminate predictions from $y_j$ that are likely contain incorrect results. 
Then, to avoid assigning multiple pseudo labels to the same object~\cite{nms}. We set an IoU filter, which eliminates out the one with the lower score among the paired bounding boxes with an IoU threshold~$\alpha_{iou}$. 
Finally, it is important to consider objects that have both true and pseudo labels.
We build a collision filter to select true sparse labels. Specifically,
we filter out the pseudo labels that have a collision exceeding the threshold~$\alpha_{col}$ with the true annotations. The above steps can be described as the following formula:
\begin{equation}
	y{'_j} = \mathop{\operatorname{Score~Filter}}(y_j, \alpha_{cls})
\end{equation}
\begin{equation}
	y{'_j} = \mathop{\operatorname{IoU~Filter}}(y{'_j}, \alpha_{iou})
\end{equation}
\begin{equation}
	y{'_j} = \mathop{\operatorname{Collision~Filter}}(y{'_j}, \alpha_{col}) \\
\end{equation}
where $y{'_j}$ is the subset of predictions after these steps. In our setting, we set a high threshold to ensure the quality of the pseudo labels.

\textbf{Prototype Label Cooperating.}  After previous steps, there may still be some positive objects that are considered negative samples due to missing detections.
To address this, we develop a practical pipeline with these steps: (1) Foreground / background separation via classification scores; (2) Identification of labeled regions (sparse labels and pseudo labels); (3) Prototype-based labeling of residual foreground areas as missed objects. Compared with MixSup~\cite{mixsup} and WS3D~\cite{weakly3d}, which also use multi-label cooperation (3D bounding boxes + BEV labels), our method reduces annotation requirements by utilizing only partial 3D bounding box annotations. This approach not only alleviates the burden of data annotation but also makes our approach more scalable while maintaining good detection performance.

\subsection{Training Strategy}\label{sec:lo}
Our method follows the standard two-stage training paradigm for limited annotations. First, we train an initial detector using sparse annotations. During this process, we introduce a prototype-based mining module. 
The loss function for the first stage is as follows:
\begin{equation}
	\mathcal{L}_{stage1} =\mathcal{L}_{det} + \mathcal{L}_{pcon} + \mathcal{L}_{pcls}
\end{equation}
where $\mathcal{L}_{det}$ is the same as the detection loss in \cite{tr3d}, and $\mathcal{L}_{pcls}$ is the Focal loss \cite{focal}. $\mathcal{L}_{pcon}$ is an Info-NCE loss~\cite{momentum}.

After obtaining the initial detection model, we use it to generate pseudo-labels for unlabeled data. In this stage, we introduce two modules: prototype-based mining and multi-label cooperative refinement. The total loss of this stage is written as:
\begin{equation}
	\mathcal{L}_{stage2} =\mathcal{L}_{stage1} + \mathcal{L}_{ref} 
\end{equation}
where $\mathcal{L}_{ref}$ is the detection loss~\cite{tr3d}, which is computed using the pseudo labels.

\definecolor{darkblue}{rgb}{0.1, 0.1, 0.5}
\definecolor{deepgreen}{rgb}{0.0, 0.6, 0.2}
\begin{table*}[t]
	\caption{Comparison with state-of-the-art methods on indoor datasets under the sparse setting. All methods are based on TR3D. }
	\label{tab:main_scannet}
	\centering
	\resizebox{0.82\textwidth}{!}{
		\begin{tabular}{c|c|c|cc|cc}
			\toprule
			\multirow{2.5}{*}{Methods}  & \multirow{2.5}{*}{Present at} & \multirow{2.5}{*}{Paradigm} & \multicolumn{2}{c|}{ScanNet V2} & \multicolumn{2}{c}{SUN RGB-D}\\ 
			\cmidrule(r){4-7}
			& &  & mAP@0.25 & mAP@0.5 & mAP@0.25 & mAP@0.5 \\
			\midrule
			 FCAF3D~\cite{fcaf3d}& ECCV & \multirow{2}{*}{Fully Supervised} & 70.7 & 56.0  & 63.8 & 48.2 \\
			TR3D~\cite{tr3d} & ICIP &  & 72.0 & 57.4 & 66.3 & 49.6 \\
			\midrule
			
			TR3D~\cite{tr3d} & ICIP & \multirow{5}{*}{Sparse Supervised} & 37.6 & 21.8& 53.9 & 36.3 \\
			Co-mining~\cite{comining} & AAAI &  &  43.3 & 26.4  & 55.9 & 39.0 \\
			SparseDet~\cite{sparsedet}  & ICCV &  &  46.0 & 28.2  & 56.7 & 38.8 \\
			CoIn~\cite{coin} & ICCV &  &  38.3 & 23.8  & 54.8 & 37.1 \\
			\rowcolor[HTML]{D0E7FF}
			CPDet3D~(ours) & - & &  \textbf{56.1}  & \textbf{40.8}  & \textbf{60.2}  & \textbf{43.3}  \\
			\bottomrule
	\end{tabular}}
\end{table*}

\begin{table*}[t]
	\caption{Comparison with state-of-the-art methods on outdoor dataset under the sparse setting. All methods are based on Voxel-RCNN. * indicates results with R11 and 3\% limited cost.}
	\label{tab:main_kitti}
	\centering
	\resizebox{0.85\textwidth}{!}{
		\begin{tabular}{c|c|c|ccc|ccc}
			\toprule
			\multirow{2.5}{*}{Methods}  & \multirow{2.5}{*}{Present at} & \multirow{2.5}{*}{Paradigm} & \multicolumn{3}{c|}{Car-3D AP (R40)} & \multicolumn{3}{c}{Car-BEV AP (R40)}\\ 
			\cmidrule(r){4-9}
			&	&  & Easy & Moderate &Hard& Easy & Moderate & Hard \\
			\midrule
			CenterPoint \cite{centerpoint} & CVPR & \multirow{2}{*}{Fully Supervised}&  89.0 & 80.5 &  76.5 & 92.9 & 89.0 & 87.5 \\ 
			Voxel-RCNN~\cite{voxelrcnn} & AAAI &  &  92.3&	85.2&	82.8&	95.5&	91.2&	88.9\\
			\midrule
			
			Voxel-RCNN~\cite{voxelrcnn}& AAAI & \multirow{5}{*}{Sparse Supervised} &  72.5 & 54.9  & 44.8 & 83.6 & 71.4 & 57.7 \\
			SS3D*~\cite{ss3d}& CVPR &  & 88.8 & 78.5  & 76.9 & - & - & - \\
			CoIn~\cite{coin} &ICCV && 84.5 & 68.4  & 58.0  & 92.3 & 81.0 & 70.2 \\
			CoIn++~\cite{coin} &ICCV && 92.0 & 79.5  & 71.5  & 96.1 & 88.8 & 82.5 \\
			\rowcolor[HTML]{D0E7FF}\
			CPDet3D~(ours) & - & &  \textbf{94.1}  &\textbf{82.2}  & \textbf{72.6}  & \textbf{96.2}  & \textbf{91.8}  & \textbf{83.9} \\
			\bottomrule
	\end{tabular}}
	\vspace {-5pt}
\end{table*}

\begin{table}
	\caption{Comparison with state-of-the-art semi supervised methods on the validation of ScanNet V2.}
	\label{tab:main_semi}
	\centering
	\resizebox{0.95\linewidth}{!}{
		\begin{tabular}{c|c|cc}
			\toprule
			Methods & Present at  & mAP@0.25 & mAP@0.5  \\
			\midrule
			SESS~\cite{sess}& CVPR & 32.0 & 14.4\\
			IoUMatch~\cite{ioumatch} & CVPR & 40.0 & 22.5  \\
			DKPE~\cite{dual}& AAAI & 44.0 &27.0   \\
			Diff-SS3D~\cite{diffusion} & NeurIPS & 43.5 &27.9   \\
			Diff3DETR~\cite{diff3detr}&ECCV &  45.1 & 29.2   \\
			DQS3D~\cite{dqs3d} & ICCV & 49.2 & 35.0   \\
			\rowcolor[HTML]{D0E7FF}
			CPDet3D~(ours)  & - & \textbf{54.6}  & \textbf{36.6}  \\
			\bottomrule
	\end{tabular}}
\vspace{-10pt}
\end{table}

\section{Experiments}\label{sec:experiment}
\subsection{Datasets and Evaluation Metrics}\label{sec:data}
We evaluate our method on two indoor datasets and one outdoor dataset: ScanNet V2~\cite{scannet}, SUN RGB-D~\cite{sunrgbd}, and KITTI~\cite{kitti}.  For indoor datasets, we use mean average precision~(mAP) with IoU thresholds of 0.25 and 0.5 as the metrics. For the outdoor dataset, we use 3D Average Precision (AP) at 40 recall thresholds (R40). 
\textbf{ScanNet V2} includes annotations for 18 object categories, with 1,201 training samples and 312 validation samples. Following the 2D method~\cite{solving}, we randomly retain one annotated object in each scene. 
\textbf{SUN RGB-D} is another indoor dataset for 3D object detection, with around 5,000 samples each for training and validation. We use one annotated object per scene for sparse supervised 3D object detection. 
\textbf{KITTI} is widely recognized for outdoor 3D object detection. We divide the 7,481 scenes into 3,712 for training and 3,769 for validation, using the same sparse supervised setting as \cite{coin} with 2\% labeled cost.


\subsection{Implementation Details}\label{sec:details}
We use TR3D~\cite{tr3d} for indoor scenes and Voxel-RCNN~\cite{voxelrcnn} with Centerhead~\cite{centerpoint} for outdoor scenes as the base detectors for our experiments. 
 All experiments are implemented using the mmdetection3d~\cite{mmdet3d} and OpenPCDet~\cite{openpcdet} frameworks, respectively. 
In terms of class-aware prototype clustering, the step number and $\kappa$ are set to 3 and 0.05. The momentum update coefficient and the prototype number for each class are set to 0.9 and 10, based on experimental results. For prototype label matching, we set the warm-up iterations and $\alpha_{pro}$ to 1000 and 0.2, respectively. In multi-label cooperative refinement module, we set $\alpha_{cls}$, $\alpha_{iou}$, and $\alpha_{col}$ to 0.2, 0.5, and 0.2.

\begin{figure*}
	\centering
	\includegraphics[width=0.95\textwidth]{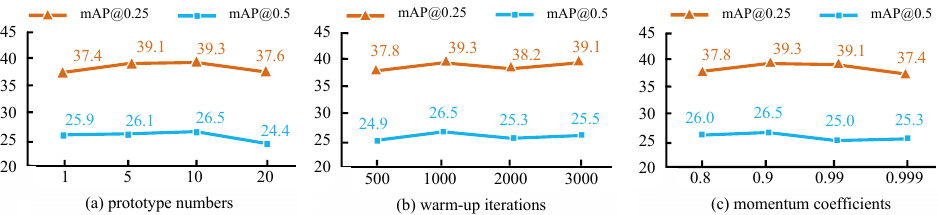}
	\caption{Results for different numbers of prototypes, warm-up iterations, and momentum coefficients.}
	\label{fig:warm_up}
	\vspace{-12pt}
\end{figure*}

\subsection{Comparisons with State-of-the-art Methods}\label{sec:vssota}
\textbf{Comparison with sparse supervised methods.} \label{sec:vsfull} 
For indoor setting, Tab.~\ref{tab:main_scannet} reports the quantitative results on the ScanNet V2~\cite{scannet} and SUN RGB-D~\cite{sunrgbd} validation sets under one object per scene setting. From Tab.~\ref{tab:main_scannet}, we can see that the performance of detectors trained on one object per sparse supervised setting decreases . For our method, we achieve average improvements of 11.3 and 4.0 compared to the SparseDet~\cite{sparsedet} on ScanNet V2 and SUN RGB-D, respectively. 
In addition, compared to fully supervised methods, our method achieves 78\% and 90\% of fully supervised performance with just one annotation per scene on the ScanNet V2 and SUN RGB-D.
For outdoor scenes, we present the results in Tab.~\ref{tab:main_kitti}.
As shown in Tab.~\ref{tab:main_kitti}, we observe a similar trend where our method outperforms CoIn++ at the moderate level, achieving a 2.7 improvement in Car-3D AP (82.2 \textit{v.s.} 79.5) and a 3.0 improvement in Car-BEV AP (91.8 \textit{v.s.} 88.8). 

\begin{figure}
	\centering
	\includegraphics[width=1.0\linewidth]{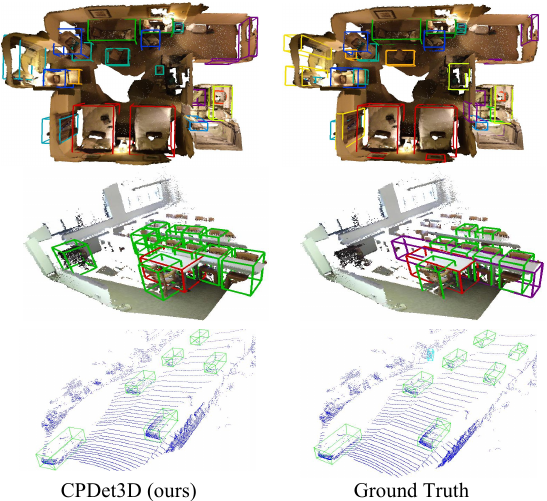}
	\caption{Visualization results of our method on the ScanNet V2, SUN RGB-D, and KITTI validation sets trained under one object per scene sparse supervised setting.}
	\label{fig:scannet_show}
	\vspace{-5pt}
\end{figure}

We present the qualitative results in Fig. \ref{fig:scannet_show}, which clearly prove the effectiveness of our approach. These results demonstrate our method is robust to point cloud scanning technologies as it extracts representations of object categories from a global perspective and enables our method to perform well across various datasets.

\begin{table}[!t]
	\caption{Statistical results regarding the recall of labels.}
	\label{tab:label_ans}
	\centering
	\resizebox{0.85\linewidth}{!}{%
		\begin{tabular}{ccccc}
			\toprule
			Number & \begin{tabular}[c]{@{}c@{}}Sparse \\ Label\end{tabular} & \begin{tabular}[c]{@{}c@{}}Prototype \\ Labels\end{tabular} & \begin{tabular}[c]{@{}c@{}}Pseudo \\ Labels\end{tabular} & mAR \\
			\midrule
			1 & \checkmark &  &  & 8.3 \\
			2 & \checkmark & \checkmark &  & 47.8  \\
			3 & \checkmark & & \checkmark & 33.4  \\
			\rowcolor[HTML]{D0E7FF}
			 4 & \checkmark & \checkmark & \checkmark & \textbf{67.1}\\
			\bottomrule
	\end{tabular}}
	\vspace{-10pt}
\end{table}

\textbf{Comparison with semi supervised methods.}\label{sec:vssemi} 
Semi supervision is another common setting for label efficient 3D object detection. To benchmark our sparse supervised method against semi supervised methods, we use the ScanNet V2~\cite{scannet}.
In this dataset of 1201 training scenes with around 20 object in each scene~\cite{scannet, dual}, a 5\% scene label rate in the semi-supervised setting (e.g., DQS3D) and one object per scene in our sparse supervised setting both yield about 1200 labeled objects, enabling a fair comparison. 
As shown in Tab.~\ref{tab:main_semi}, our model achieves an advantage over the SOTA semi supervised method DQS3D on ScanNet V2, \textit{i.e.}, 5.4 on mAP@0.25 and 1.6 on mAP@0.5.

\subsection{Label Statistical Analysis}\label{sec:stats}
The quality of mined labels determines the upper bound of detection performance. Therefore, we analyzed the quality of mined labels on the ScanNet V2 training set in terms of both precision and recall.
In terms of precision, pseudo labels defined by overlap with ground truth have a precision of 95.5, ensuring that most contribute positively to model performance. Prototype labels, defined within the ground truth bounding box, have a precision of about 71.1. During inference, Non-Maximum Suppression (NMS) filters low-quality predictions, reducing the impact of errors.

In terms of recall, we use mean average recall~(mAR) as the metric. From Tab. \ref{tab:label_ans}, we can see that only sparse labels are used, and the mAR is 8.3. Using sparse and prototype labels results in an mAR of 47.8, which is a slight improvement. In contrast, using sparse and pseudo labels alone increases the mAR significantly to 33.4. When all labels are combined, the mAR further increases to 67.1, confirming that the complementarity between these two types of labels provides  more useful labels.

\begin{table}[!t]
	\captionof{table}{Ablation study of key components including PLM, CPC, and MCR.}
	\label{tab:components}
	\centering
	\resizebox{1\linewidth}{!}{%
		\begin{tabular}{cccc|cc}
			\toprule
			Number & PLM & CPC & MCR & \text{mAP}@0.25 & \text{mAP}@0.5 \\
			\midrule
			1 &  & & & 37.6 & 21.8\\
			2 & \checkmark &  & & 38.0 & 23.5  \\
			3 & \checkmark & \checkmark & & 39.3 & 26.5  \\
			4 &  & & \checkmark & 51.9 & 32.3 \\
			5 & \checkmark &  & \checkmark & 54.2 & 38.3  \\
			\rowcolor[HTML]{D0E7FF}
			6 & \checkmark & \checkmark & \checkmark & \textbf{56.1} & \textbf{40.8} \\
			\bottomrule
	\end{tabular}}
	\vspace{-5pt}
\end{table}

\subsection{Ablation Study}\label{sec:ablation}
We conduct extensive ablation studies on the one object per scene setting of ScanNet V2. 

\textbf{Effect of Different Components.} \label{sec:components} 
We first ablate the effects of different components of our model in Tab. \ref{tab:components}. The base model is TR3D~\cite{tr3d} under the one object per scene setting.
Introducing prototype label matching (``PLM'') in experiment 2 and just using the classification score for label matching slightly improves mAP@0.25 from 37.6 to 38.0. This result shows that label assignment based solely on the classification score is suboptimal in a sparse supervised setting.
Next, comparing experiment 2 and experiment 3, adding the class-aware prototype clustering (``CPC'') boosts mAP@0.25 from 38.0 to 39.3 and mAP@0.5 from 23.6 to 26.5. This variant proves that simultaneously considering the classification score and feature similarity can provide more accurate prototype labels.
Finally, with the multi-label cooperative refinement module (``MCR'') in experiment 6 achieves the best results, highlighting its benefits.

We also evaluate additional component combinations, including MCR alone and the combination of PLM and MCR in Tab. \ref{tab:components}. CPC alone is not evaluated, as it does not directly impact performance. For MCR alone, we remove prototype labels and use only sparse and pseudo labels. Comparing experiment 1 and experiment 4 shows significant improvement, confirming the effectiveness of iterative training. Adding PLM in experiment 5 further boosts mAP@0.25 and mAP@0.5 from 51.9 to 54.2 and from 32.3 to 38.3, proving that prototype labels help recover missed detections in iterative training.

\begin{table}
	\caption{$\alpha_{pro}$ and $\alpha_{cls}$ at different thresholds for mAP@0.5.}
	\label{tab:alpha}
	\centering
	\resizebox{\linewidth}{!}{
		\begin{tabular}{c|ccccccc}
			\toprule
			Threshold & 0.01 & 0.1 & 0.2 & 0.3 & 0.4 & 0.5 & 0.6 \\
			\midrule
			$\alpha_{pro}$ & 17.3 & 24.4 & \cellcolor[HTML]{D0E7FF}\textbf{26.5} & 24.1 & 21.8 & 21.8 & 21.8\\
			$\alpha_{cls}$ & 33.7 & 39.5 & \cellcolor[HTML]{D0E7FF}\textbf{40.8} & 38.1 & 36.7 & 33.0 & 32.3\\
			\bottomrule
	\end{tabular}}
	\vspace{-10pt}
\end{table}

\textbf{Choices in Prototype-based Object Mining.} \label{sec:parameterscpc}
In this section, we study the impact of the number of prototypes, the iterations of warm-up, and the momentum coefficient. Fig. \ref{fig:warm_up} (a) suggests that using 10 prototypes for each class is most effective for representing the differences among objects in the same category. The different choices of warm-up iterations and momentum coefficients are shown in Fig. \ref{fig:warm_up} (b) and (c), which indicate that with an iteration of 1000 and a coefficient of 0.9, our method achieves the best performance. This is because a suitable warm-up iteration and momentum coefficient can make the generated prototype labels more stable, thus improving the performance. Besides, we conduct experiments to compare performance at different $\alpha_{pro}$ score thresholds in Tab. \ref{tab:alpha}. Since $\alpha_{pro} = 0.2$ yields the best result, we use this value as the default.

\begin{figure}[t]
	\centering
	\includegraphics[width=0.9\linewidth]{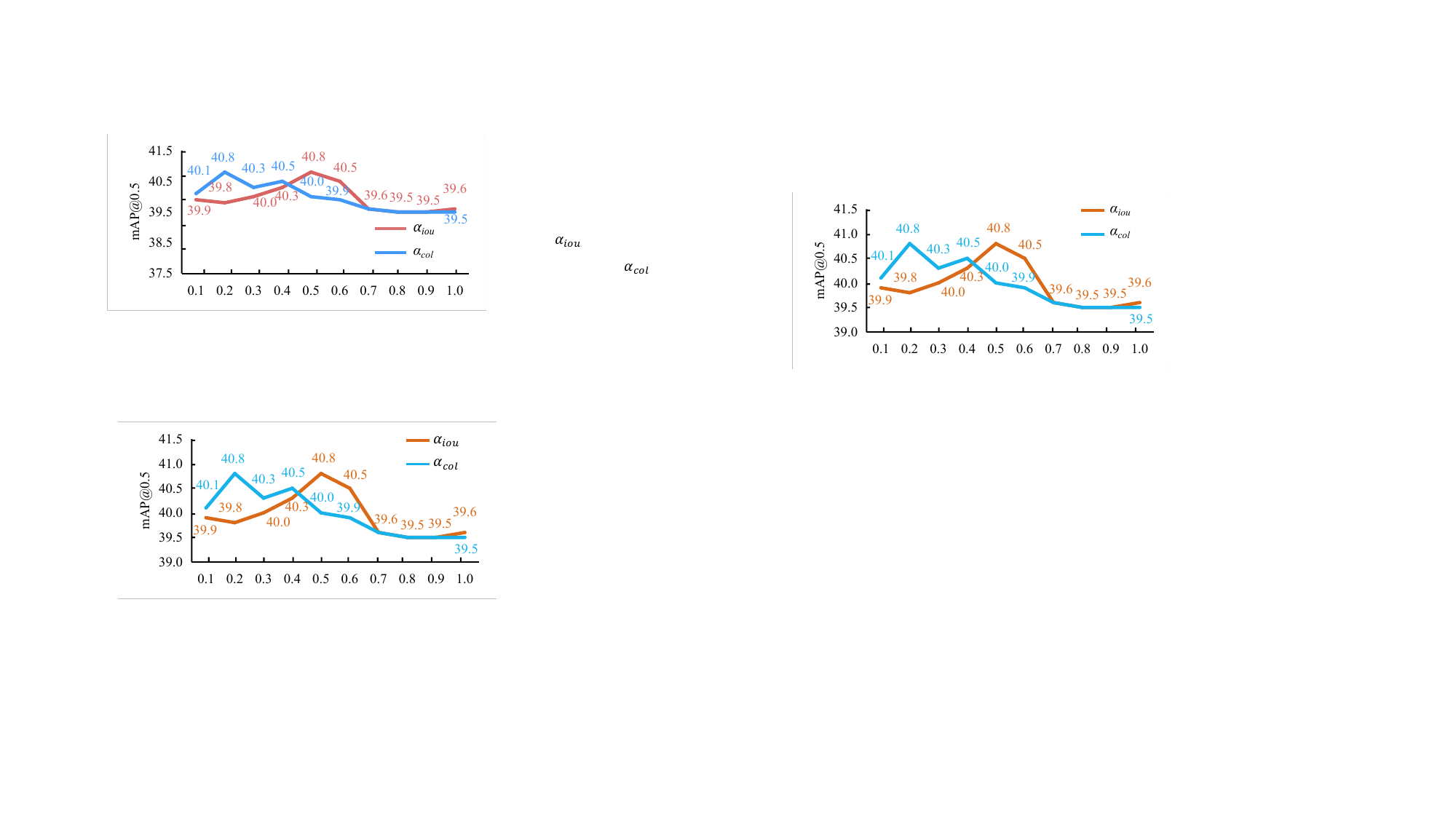}
	\caption{Ablation study of $\alpha_{iou}$ and $\alpha_{col}$.}
	\label{fig:iou_col}
	\vspace{-10pt}
\end{figure} 
\textbf{Impact of Multi-label Cooperative Refinement.}\label{sec:parametersmc}
In this section, we study the impact of hyper-parameters selection of multi-label cooperative refinement module. The hyper-parameters include classification score threshold $\alpha_{cls}$, IoU threshold $\alpha_{iou}$, and collision threshold $\alpha_{col}$. We adjust the $\alpha_{cls}$ from 0.01 to 0.6 to conduct ablation studies.  
As shown in Tab. \ref{tab:alpha}, the accuracy at $\alpha_{cls}$ = 0.2 is much better than the other choices, thus we choose $\alpha_{cls}$ = 0.2. 
From Fig. \ref{fig:iou_col}, the optimal results for the parameters $\alpha_{iou}$ and $\alpha_{col}$ are observed at $\alpha_{iou}$ = 0.5 and $\alpha_{col}$ = 0.2, where the mAP@0.5 reaches its highest point. 
The figure also demonstrates that variations in these parameters result in only minor fluctuations in performance. 

\section{Conclusion}\label{sec:conclusion}
In this paper, we proposed a unified sparse supervised 3D object detection method for indoor and outdoor environments. We presented a prototype-based object mining module with class-aware prototype clustering to learn class prototypes from limited labeled objects and prototype label matching to assign labels to unlabeled objects based on learned similarity. Additionally, a multi-label cooperative refinement module is designed to integrate high-quality pseudo labels with prototype labels, avoiding complex iterative design. Experiments show that our method achieves 78\%, 90\%, and 96\% of the performance of the fully supervised detector in the one object per scene sparse supervised setting on ScanNet V2, SUN RGB-D, and KITTI.

\textbf{Acknowledgments.} This work was supported by the National Key R\&D Program of China No. 2024YFC3015801,  National Science Fund of China (Grant Nos. 62361166670, 62276144, U24A20330, 62306238 and 62176124) and the Fundamental Research Funds for the Central Universities.

{
	\small
	\bibliographystyle{ieeenat_fullname}
	\bibliography{main}

\begin{thebibliography}{56}
\providecommand{\natexlab}[1]{#1}
\providecommand{\url}[1]{\texttt{#1}}
\expandafter\ifx\csname urlstyle\endcsname\relax
  \providecommand{\doi}[1]{doi: #1}\else
  \providecommand{\doi}{doi: \begingroup \urlstyle{rm}\Url}\fi

\bibitem[Chen et~al.(2022)Chen, Li, Chen, Wang, Zhang, and Hua]{denselearn}
Binghui Chen, Pengyu Li, Xiang Chen, Biao Wang, Lei Zhang, and Xian-Sheng Hua.
\newblock {Dense Learning based Semi-Supervised Object Detection}.
\newblock In \emph{CVPR}, 2022.

\bibitem[Contributors(2020)]{mmdet3d}
MMDetection3D Contributors.
\newblock {MMDetection3D: OpenMMLab} next-generation platform for general {3D} object detection.
\newblock \url{https://github.com/open-mmlab/mmdetection3d}, 2020.

\bibitem[Cuturi(2013)]{sinkhorn}
Marco Cuturi.
\newblock {Sinkhorn Distances: Lightspeed Computation of Optimal Transport}.
\newblock In \emph{NeurIPS}, 2013.

\bibitem[Dai et~al.(2017)Dai, Chang, Savva, Halber, Funkhouser, and Nie{\ss}ner]{scannet}
Angela Dai, Angel~X. Chang, Manolis Savva, Maciej Halber, Thomas Funkhouser, and Matthias Nie{\ss}ner.
\newblock {{ScanNet}: Richly-Annotated 3{D} Reconstructions of Indoor Scenes}.
\newblock In \emph{CVPR}, 2017.

\bibitem[Deng et~al.(2021)Deng, Shi, Li, Zhou, Zhang, and Li]{voxelrcnn}
Jiajun Deng, Shaoshuai Shi, Peiwei Li, Wengang Zhou, Yanyong Zhang, and Houqiang Li.
\newblock {Voxel R-CNN: Towards High Performance Voxel-based 3D Object Detection}.
\newblock In \emph{AAAI}, 2021.

\bibitem[Deng et~al.(2024)Deng, Lu, and Zhang]{diff3detr}
Jiacheng Deng, Jiahao Lu, and Tianzhu Zhang.
\newblock {Diff3DETR: Agent-based Diffusion Model for Semi-supervised 3D Object Detection}.
\newblock In \emph{ECCV}, 2024.

\bibitem[Gao et~al.(2023)Gao, Tian, Li, Zhao, and Zhou]{dqs3d}
Huan-ang Gao, Beiwen Tian, Pengfei Li, Hao Zhao, and Guyue Zhou.
\newblock {{DQS3D}: Densely-matched Quantization-aware Semi-supervised 3{D} Detection}.
\newblock In \emph{ICCV}, 2023.

\bibitem[Geiger et~al.(2012)Geiger, Lenz, and Urtasun]{kitti}
Andreas Geiger, Philip Lenz, and Raquel Urtasun.
\newblock {Are we ready for autonomous driving? The KITTI vision benchmark suite}.
\newblock In \emph{CVPR}, 2012.

\bibitem[Han et~al.(2024)Han, Zhao, Chen, Ma, and Zhang]{dual}
Yucheng Han, Na Zhao, Weiling Chen, Keng~Teck Ma, and Hanwang Zhang.
\newblock {Dual-Perspective Knowledge Enrichment for Semi-Supervised 3{D} Object Detection}.
\newblock In \emph{AAAI}, 2024.

\bibitem[He et~al.(2020)He, Fan, Wu, Xie, and Girshick]{momentum}
Kaiming He, Haoqi Fan, Yuxin Wu, Saining Xie, and Ross Girshick.
\newblock {Momentum Contrast for Unsupervised Visual Representation Learning}.
\newblock In \emph{CVPR}, 2020.

\bibitem[Ho et~al.(2024)Ho, Tai, Lin, Yang, and Tsai]{diffusion}
Cheng-Ju Ho, Chen-Hsuan Tai, Yen-Yu Lin, Ming-Hsuan Yang, and Yi-Hsuan Tsai.
\newblock {Diffusion-SS3D: Diffusion Model for Semi-supervised 3D Object Detection}.
\newblock In \emph{NeurIPS}, 2024.

\bibitem[Hosang et~al.(2017)Hosang, Benenson, and Schiele]{nms}
J. Hosang, R. Benenson, and B. Schiele.
\newblock {Learning Non-Maximum Suppression}.
\newblock In \emph{CVPR}, 2017.

\bibitem[Hui et~al.(2021)Hui, Yuan, Cheng, Xie, Zhang, and Yang]{hui2021superpoint}
Le Hui, Jia Yuan, Mingmei Cheng, Jin Xie, Xiaoya Zhang, and Jian Yang.
\newblock {Superpoint Network for Point Cloud Oversegmentation}.
\newblock In \emph{ICCV}, 2021.

\bibitem[Hui et~al.(2022{\natexlab{a}})Hui, Tang, Shen, Xie, and Yang]{hui2022learning}
Le Hui, Linghua Tang, Yaqi Shen, Jin Xie, and Jian Yang.
\newblock {Learning Superpoint Graph Cut for 3D Instance Segmentation}.
\newblock In \emph{NeurIPS}, 2022{\natexlab{a}}.

\bibitem[Hui et~al.(2022{\natexlab{b}})Hui, Wang, Tang, Lan, Xie, and Yang]{hui20223d}
Le Hui, Lingpeng Wang, Linghua Tang, Kaihao Lan, Jin Xie, and Jian Yang.
\newblock {3D Siamese Transformer Network for Single Object Tracking on Point Clouds}.
\newblock In \emph{ECCV}, 2022{\natexlab{b}}.

\bibitem[Lang et~al.(2019)Lang, Vora, Caesar, Zhou, Yang, and Beijbom]{pointpillars}
Alex~H Lang, Sourabh Vora, Holger Caesar, Lubing Zhou, Jiong Yang, and Oscar Beijbom.
\newblock {PointPillars: Fast Encoders for Object Detection from Point Clouds}.
\newblock In \emph{CVPR}, 2019.

\bibitem[Li et~al.(2022{\natexlab{a}})Li, Li, Wang, Yichao, Liang, and Zhang]{denseteacher}
Gang Li, Xiang Li, Yujie Wang, Wu Yichao, Ding Liang, and Shanshan Zhang.
\newblock {DTG-SSOD: Dense Teacher Guidance for Semi-Supervised Object Detection}.
\newblock In \emph{NeurIPS}, 2022{\natexlab{a}}.

\bibitem[Li et~al.(2022{\natexlab{b}})Li, Pan, Yan, Tang, and Zheng]{siod}
Hanjun Li, Xingjia Pan, Ke Yan, Fan Tang, and Wei-Shi Zheng.
\newblock {SIOD: Single Instance Annotated Per Category Per Image for Object Detection}.
\newblock In \emph{CVPR}, 2022{\natexlab{b}}.

\bibitem[Lin et~al.(2017)Lin, Goyal, Girshick, He, and Doll{\'a}r]{focal}
Tsung-Yi Lin, Priya Goyal, Ross Girshick, Kaiming He, and Piotr Doll{\'a}r.
\newblock {Focal Loss for Dense Object Detection}.
\newblock In \emph{ICCV}, 2017.

\bibitem[Liu et~al.(2022)Liu, Gao, Liu, Liu, Meng, and Gao]{ss3d}
Chuandong Liu, Chenqiang Gao, Fangcen Liu, Jiang Liu, Deyu Meng, and Xinbo Gao.
\newblock {{SS3D}: Sparsely-Supervised 3{D} Object Detection from Point Cloud}.
\newblock In \emph{CVPR}, 2022.

\bibitem[Liu et~al.(2025)Liu, Liu, Li, Ren, and Xu]{liu2025milnet}
Jinfu Liu, Hong Liu, Xia Li, Jiale Ren, and Xinhua Xu.
\newblock {MiLNet: Multiplex Interactive Learning Network for RGB-T Semantic Segmentation}.
\newblock \emph{TIP}, 2025.

\bibitem[Lv et~al.(2024{\natexlab{a}})Lv, Liu, and Li]{lv2024context}
Ying Lv, Zhi Liu, and Gongyang Li.
\newblock {Context-Aware Interaction Network for RGB-T Semantic Segmentation}.
\newblock \emph{TMM}, pages 6348--6360, 2024{\natexlab{a}}.

\bibitem[Lv et~al.(2024{\natexlab{b}})Lv, Liu, Li, and Chang]{lv2024noise}
Ying Lv, Zhi Liu, Gongyang Li, and Xiaojun Chang.
\newblock {Noise-Aware Intermediary Fusion Network For Off-Road Freespace Detection}.
\newblock \emph{TIV}, 2024{\natexlab{b}}.

\bibitem[Meng et~al.(2020)Meng, Wang, Zhou, Shen, Van~Gool, and Dai]{weakly3d}
Qinghao Meng, Wenguan Wang, Tianfei Zhou, Jianbing Shen, Luc Van~Gool, and Dengxin Dai.
\newblock {Weakly Supervised 3{D} Object Detection from Lidar Point Cloud}.
\newblock In \emph{ECCV}, 2020.

\bibitem[Qi et~al.(2019)Qi, Litany, He, and Guibas]{votenet}
Charles~R Qi, Or Litany, Kaiming He, and Leonidas~J Guibas.
\newblock {Deep Hough Voting for 3{D} Object Detection in Point Clouds}.
\newblock In \emph{ICCV}, 2019.

\bibitem[Rambhatla et~al.(2023)Rambhatla, Suri, Chellappa, and Shrivastava]{sparsedet}
Sai~Saketh Rambhatla, Saksham Suri, Rama Chellappa, and Abhinav Shrivastava.
\newblock {SparseDet: Improving Sparsely Annotated Object Detection with Pseudo-positive Mining}.
\newblock In \emph{ICCV}, 2023.

\bibitem[Rukhovich et~al.(2022)Rukhovich, Vorontsova, and Konushin]{fcaf3d}
Anna Rukhovich, Anna Vorontsova, and Anton Konushin.
\newblock {FCAF3D: Fully Convolutional Anchor-Free 3{D} Object Detection}.
\newblock In \emph{ECCV}, 2022.

\bibitem[Rukhovich et~al.(2023)Rukhovich, Vorontsova, and Konushin]{tr3d}
Danila Rukhovich, Anna Vorontsova, and Anton Konushin.
\newblock {{TR3D}: Towards Real-Time Indoor 3{D} Object Detection}.
\newblock In \emph{ICIP}, 2023.

\bibitem[Shi et~al.(2022)Shi, Li, and Ma]{pillarnet}
Guangsheng Shi, Ruifeng Li, and Chao Ma.
\newblock {PillarNet: Real-Time and High-Performance Pillar-based 3D Object Detection}.
\newblock In \emph{ECCV}, 2022.

\bibitem[Song et~al.(2015)Song, Lichtenberg, and Xiao]{sunrgbd}
Shuran Song, Samuel~P. Lichtenberg, and Jianxiong Xiao.
\newblock {{SUN RGB-D}: A RGB-D Scene Understanding Benchmark Suite}.
\newblock In \emph{CVPR}, 2015.

\bibitem[Tang et~al.(2022)Tang, Hui, and Xie]{tang2022learning}
Linghua Tang, Le Hui, and Jin Xie.
\newblock {Learning Inter-Superpoint Affinity for Weakly Supervised 3D Instance Segmentation}.
\newblock In \emph{ACCV}, 2022.

\bibitem[Tarvainen and Valpola(2017)]{meanteacher}
Antti Tarvainen and Harri Valpola.
\newblock {Mean teachers are better role models: Weight-averaged consistency targets improve semi-supervised deep learning results}.
\newblock In \emph{NeurIPS}, 2017.

\bibitem[Team(2020)]{openpcdet}
OpenPCDet~Development Team.
\newblock Openpcdet: An open-source toolbox for 3d object detection from point clouds.
\newblock \url{https://github.com/open-mmlab/OpenPCDet}, 2020.

\bibitem[Wang et~al.(2021{\natexlab{a}})Wang, Cong, Litany, Gao, and Guibas]{ioumatch}
He Wang, Yezhen Cong, Or Litany, Yue Gao, and Leonidas~J Guibas.
\newblock {{3DIoUMatch}: Leveraging IoU Prediction for Semi-Supervised 3{D} Object Detection}.
\newblock In \emph{CVPR}, 2021{\natexlab{a}}.

\bibitem[Wang et~al.(2022{\natexlab{a}})Wang, Dong, Shi, Li, Li, Li, Wang, et~al.]{cagroup}
Haiyang Wang, Shaocong Dong, Shaoshuai Shi, Aoxue Li, Jianan Li, Zhenguo Li, Liwei Wang, et~al.
\newblock {CAGroup3D: Class-Aware Grouping for 3{D} Object Detection on Point Clouds}.
\newblock In \emph{NeurIPS}, 2022{\natexlab{a}}.

\bibitem[Wang et~al.(2022{\natexlab{b}})Wang, Shi, Yang, Fang, Qian, Li, Schiele, and Wang]{rgbnet}
Haiyang Wang, Shaoshuai Shi, Ze Yang, Rongyao Fang, Qi Qian, Hongsheng Li, Bernt Schiele, and Liwei Wang.
\newblock {{RBGNet}: Ray-Based Grouping for 3{D} Object Detection}.
\newblock In \emph{CVPR}, 2022{\natexlab{b}}.

\bibitem[Wang et~al.(2023)Wang, Liu, Zhang, Zhang, Zhang, Gan, Wang, Wang, and Wang]{calibrated}
Haohan Wang, Liang Liu, Boshen Zhang, Jiangning Zhang, Wuhao Zhang, Zhenye Gan, Yabiao Wang, Chengjie Wang, and Haoqian Wang.
\newblock {Calibrated Teacher for Sparsely Annotated Object Detection}.
\newblock In \emph{AAAI}, 2023.

\bibitem[Wang et~al.(2021{\natexlab{b}})Wang, Yang, Cao, and Zhang]{comining}
Tiancai Wang, Tong Yang, Jiale Cao, and Xiangyu Zhang.
\newblock {Co-mining: Self-Supervised Learning for Sparsely Annotated Object Detection}.
\newblock In \emph{AAAI}, 2021{\natexlab{b}}.

\bibitem[Wang et~al.(2021{\natexlab{c}})Wang, Zhou, Yu, Dai, Konukoglu, and Van~Gool]{exploring}
Wenguan Wang, Tianfei Zhou, Fisher Yu, Jifeng Dai, Ender Konukoglu, and Luc Van~Gool.
\newblock {Exploring Cross-Image Pixel Contrast for Semantic Segmentation}.
\newblock In \emph{ICCV}, 2021{\natexlab{c}}.

\bibitem[Wang et~al.(2022{\natexlab{c}})Wang, Han, Zhou, and Liu]{visual}
Wenguan Wang, Cheng Han, Tianfei Zhou, and Dongfang Liu.
\newblock {Visual Recognition with Deep Nearest Centroids}.
\newblock \emph{arXiv preprint arXiv:2209.07383}, 2022{\natexlab{c}}.

\bibitem[Wu et~al.(2024)Wu, Zhang, Qian, Xie, and Yang]{wu2024text2lidar}
Yang Wu, Kaihua Zhang, Jianjun Qian, Jin Xie, and Jian Yang.
\newblock {Text2lidar: Text-Guided Lidar Point Cloud Generation via Equirectangular Transformer}.
\newblock In \emph{ECCV}, 2024.

\bibitem[Xia et~al.(2023)Xia, Deng, Wen, Wu, Shi, Li, and Wang]{coin}
Qiming Xia, Jinhao Deng, Chenglu Wen, Hai Wu, Shaoshuai Shi, Xin Li, and Cheng Wang.
\newblock {{CoIn}: Contrastive Instance Feature Mining for Outdoor 3{D} Object Detection with Very Limited Annotations}.
\newblock In \emph{ICCV}, 2023.

\bibitem[Xia et~al.(2024)Xia, Ye, Wu, Zhao, Xing, Huang, Deng, Li, Wen, and Wang]{hinted}
Qiming Xia, Wei Ye, Hai Wu, Shijia Zhao, Leyuan Xing, Xun Huang, Jinhao Deng, Xin Li, Chenglu Wen, and Cheng Wang.
\newblock {HINTED: Hard Instance Enhanced Detector with Mixed-Density Feature Fusion for Sparsely-Supervised 3D Object Detection}.
\newblock In \emph{CVPR}, 2024.

\bibitem[Xu et~al.(2022)Xu, Wang, Zheng, Rao, Zhou, and Lu]{backtoreal}
Xiuwei Xu, Yifan Wang, Yu Zheng, Yongming Rao, Jie Zhou, and Jiwen Lu.
\newblock {Back to Reality: Weakly-supervised 3D Object Detection with Shape-guided Label Enhancement}.
\newblock In \emph{CVPR}, 2022.

\bibitem[Yan et~al.(2018)Yan, Mao, and Li]{second}
Yan Yan, Yuxing Mao, and Bo Li.
\newblock {SECOND: Sparsely Embedded Convolutional Detection}.
\newblock \emph{Sensors}, 18\penalty0 (10):\penalty0 3337, 2018.

\bibitem[Yang et~al.(2024)Yang, Fan, and Zhang]{mixsup}
Yuxue Yang, Lue Fan, and Zhaoxiang Zhang.
\newblock Mixsup: Mixed-grained supervision for label-efficient lidar-based 3d object detection.
\newblock \emph{arXiv preprint arXiv:2401.16305}, 2024.

\bibitem[Yin et~al.(2021)Yin, Zhou, and Krahenbuhl]{centerpoint}
Tianwei Yin, Xingyi Zhou, and Philipp Krahenbuhl.
\newblock {Center-based 3D Object Detection and Tracking}.
\newblock In \emph{CVPR}, 2021.

\bibitem[Zhang et~al.(2024{\natexlab{a}})Zhang, Zhou, Li, He, Ma, Zhang, Yao, Zhang, and Wang]{uncovering}
Fei Zhang, Tianfei Zhou, Boyang Li, Hao He, Chaofan Ma, Tianjiao Zhang, Jiangchao Yao, Ya Zhang, and Yanfeng Wang.
\newblock {Uncovering Prototypical Knowledge for Weakly Open-Vocabulary Semantic Segmentation}.
\newblock In \emph{NeurIPS}, 2024{\natexlab{a}}.

\bibitem[Zhang et~al.(2020)Zhang, Chen, Shen, Hao, Zhu, and Savvides]{solving}
Han Zhang, Fangyi Chen, Zhiqiang Shen, Qiqi Hao, Chenchen Zhu, and Marios Savvides.
\newblock {Solving Missing-Annotation Object Detection with Background Recalibration Loss}.
\newblock In \emph{ICASSP}, 2020.

\bibitem[Zhang et~al.(2024{\natexlab{b}})Zhang, Hui, Li, and Dai]{zhang20243d}
Liyuan Zhang, Le Hui, Bo Li, and Yuchao Dai.
\newblock {3D Focusing-and-Matching Network for Multi-Instance Point Cloud Registration}.
\newblock In \emph{NeurIPS}, 2024{\natexlab{b}}.

\bibitem[Zhao et~al.(2020)Zhao, Chua, and Lee]{sess}
Na Zhao, Tat-Seng Chua, and Gim~Hee Lee.
\newblock {{SESS}: Self-Ensembling Semi-Supervised 3{D} Object Detection}.
\newblock In \emph{CVPR}, 2020.

\bibitem[Zhao and Qi(2022)]{prototypical}
Shizhen Zhao and Xiaojuan Qi.
\newblock {Prototypical VoteNet for Few-Shot 3D Point Cloud Object Detection}.
\newblock In \emph{NeurIPS}, 2022.

\bibitem[Zhou et~al.(2022{\natexlab{a}})Zhou, Ge, Liu, Mao, Li, Yu, and Sun]{densepse}
Hongyu Zhou, Zheng Ge, Songtao Liu, Weixin Mao, Zeming Li, Haiyan Yu, and Jian Sun.
\newblock {Dense Teacher: Dense Pseudo-Labels for Semi-supervised Object Detection}.
\newblock In \emph{ECCV}, 2022{\natexlab{a}}.

\bibitem[Zhou et~al.(2022{\natexlab{b}})Zhou, Wang, Konukoglu, and Van~Gool]{rethinkseg}
Tianfei Zhou, Wenguan Wang, Ender Konukoglu, and Luc Van~Gool.
\newblock {Rethinking Semantic Segmentation: A Prototype View}.
\newblock In \emph{CVPR}, 2022{\natexlab{b}}.

\bibitem[Zhou and Tuzel(2018)]{voxelnet}
Yin Zhou and Oncel Tuzel.
\newblock {VoxelNet: End-to-End Learning for Point Cloud Based 3{D} Object Detection}.
\newblock In \emph{CVPR}, 2018.

\bibitem[Zhu et~al.(2024)Zhu, Hui, Shen, and Xie]{spgroup3d}
Yun Zhu, Le Hui, Yaqi Shen, and Jin Xie.
\newblock {{SPGroup3D}: Superpoint Grouping Network for Indoor 3{D} Object Detection}.
\newblock In \emph{AAAI}, 2024.

\end{thebibliography}
}

\clearpage
\setcounter{page}{1}
\maketitlesupplementary

\section{Overview}
\label{sec:intor}
In this supplementary material, we first analyze the effect of semantic and bounding box prediction under limited annotations (\S \ref{sec:clsreg}) and examine the distribution of our labeled splits by category on indoor datasets (\S \ref{sec:annotation}). Then, we provide additional experiments (\S \ref{sec:more_exp}), including the performance of our model under different annotation settings (\S \ref{sec:more_sparse}), an evaluation of our method extended to other detectors (\S \ref{sec:more_det}) and comparison between prototypes and GT sampling \ref{sec:pro_gt}), indoor per-category evaluation results (\ref{sec:per_class}), and further qualitative visualizations (\S \ref{sec:visual}).  Finally, we discuss the limitation and further work (\S \ref{sec:limitations}) of our proposed method.

\begin{table}[H]
	\vspace{-5pt}
	\captionof{table}{Ablation study of the importance of class and regress branches based on ScanNet V2.}
	\centering
	\label{tab:clsreg}
	\resizebox{1.0\linewidth}{!}{\begin{tabular}{ccc|cc}
			\toprule
			Number & Class & Regress &  \text{mAP}@0.25 & \text{mAP}@0.5   \\
			\midrule
			1 & &  & 37.9 & 21.7\\
			2 & \checkmark &  & 50.1 & 31.5\\
			3 & \checkmark & \checkmark  & \textbf{51.9} & \textbf{32.3}   \\
			\bottomrule
\end{tabular}}
\vspace{-10pt}
\end{table}
\section{More Analysis}\label{sec:more_ans}  
\subsection{Effect Analysis}\label{sec:clsreg}  
In the prototype-based object mining module, we only generate the category labels based on prototypes without generating box labels. To further understand the relative importance of semantic versus bounding box prediction under conditions of limited labeling, we design an experiment on ScanNet V2~\cite{scannet}. For simplicity, we remove the prototype-based object mining module and simplify the multi-label collaborative refinement module to use sparse and pseudo labels. To make a fair comparison, we also ensure that other conditions are consistent, such as thresholds. This allows us to isolate and evaluate the contributions of semantic learning independently based on whether to use the pseudo labels' boxes as regression targets. 

As detailed in the results presented in the second and third row of Tab. \ref{tab:clsreg}, we observe a modest decrease in detection performance when only the class branch of the multi-label cooperative refinement module is active. Specifically, there are declines of 1.8 and 0.8 on mAP@0.25 and mAP@0.5. This relatively small reduction in performance suggests that semantic understanding plays a more important role than bounding box localization in 3D object detection with sparse annotations. We think the reason for this discovery is that the 3D point cloud essentially provides extensive geometric and structural information about the environment, as well as the objects within it.

\subsection{Annotation Analysis}\label{sec:annotation}  
Unlike the well-established sparse 3D object detection methods for outdoor environments~\cite{ss3d, coin}, indoor counterparts remain in a start stage, hindered by the absence of readily available annotated splits. This lack of sufficient tailored data has significantly slowed progress in indoor sparse 3D object detection. To address this limitation, we construct two dataset splits based on the widely recognized ScanNet V2~\cite{scannet} and SUN RGB-D~\cite{sunrgbd} datasets to advance research in this domain.

\begin{figure*}
	\centering
	\includegraphics[width=1.0\textwidth]{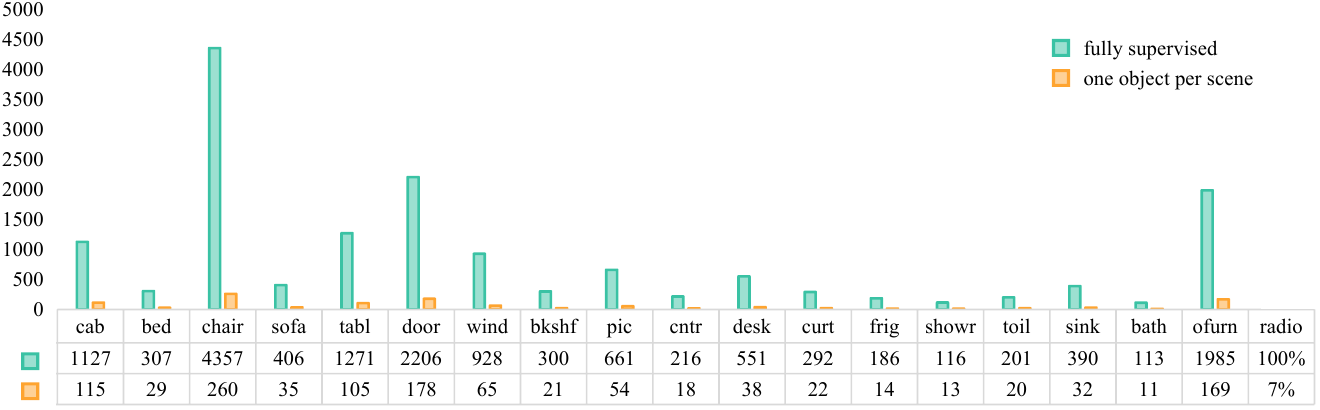}
	\caption{Per category analysis of the number of labels on ScanNet V2.}
	\label{fig:scannet_data_show}
\end{figure*}

\begin{figure*}
	\centering
	\includegraphics[width=0.7\textwidth]{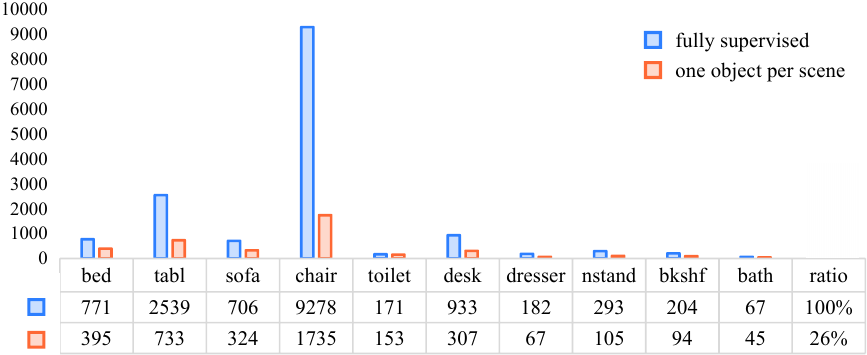}
	\caption{Per category analysis of the number of labels on SUN RGB-D.}
	\label{fig:sunrgbd_data_show}
	\vspace{-10pt}
\end{figure*}

We provide a detailed analysis of the dataset splits, including the precise number of labeled instances for each object category, as illustrated in Fig. \ref{fig:scannet_data_show} and Fig. \ref{fig:sunrgbd_data_show}. Our analysis reveals a noticeable class imbalance in both datasets under the fully supervised setting. Specifically, the most frequently occurring category, ``chair'', exhibits the highest number of annotations, while the ``bathtub'' category is significantly underrepresented, highlighting a disparity in data distribution. Furthermore, when analyzing the setting where each scene contains only one object per category, we observe that the issue of class imbalance is notably mitigated.

\section{More Experiments} \label{sec:more_exp}
\begin{table}
	\caption{The results on validation of ScanNet V2 trained on fully and sparse supervised settings on the validate of ScanNet V2.}
	\label{tab:ocoa_scannet}
	\centering
	\resizebox{1.0\linewidth}{!}{
		\begin{tabular}{c|c|cc}
			\toprule
			Methods & Paradigm & \text{mAP}@0.25 & \text{mAP}@0.5\\ 
			\midrule
			TR3D~\cite{tr3d} & Fully Supervised  & 72.0 & 57.4  \\
			\midrule
			TR3D~\cite{tr3d}  & Two Objects &  43.5 & 30.3   \\
			ours  & Per Scene &  \textbf{60.7} & \textbf{45.5}  \\
			\midrule
			TR3D~\cite{tr3d} & Three Objects &  48.3 & 34.2   \\
			ours &  Per Scene &  \textbf{64.1} & \textbf{47.7}  \\
			\midrule
			TR3D~\cite{tr3d}  &  One Object &  62.9 & 48.5   \\
			ours  & Per Class Per Scene  &  \textbf{69.6} & \textbf{55.5}  \\
			\bottomrule
	\end{tabular}}
			\vspace{-10pt}
\end{table}
\subsection{More Sparse Supervised Settings} \label{sec:more_sparse}
In this section, we first add the labeled objects in each scene on ScanNet V2~\cite{scannet},  \textit{i.e.}, 2 objects per scene and 3 objects per scene, to verify the effectiveness of our model in more labeled cases. Tab. \ref{tab:ocoa_scannet} indicates that our model is extensible in more labeled cases, but the results that can be improved are gradually reduced because the best case of our model is fully supervised performance. In the case of 2 annotations, our approach achieves 82\% of fully supervised performance. In addition, 87\% of fully supervised performance is achieved in the three labeled objects setting. 

Recently, apart from random annotation, another form of sparse annotation has also attracted researchers' attention in 2D object detection, which is called one object per class per scene~\cite{siod}. This paradigm, by ensuring at least one instance of each category is labeled, not only enables a more comprehensive perception of all categories present in the scene but also generates a more diverse set of annotations. From Tab. \ref{tab:ocoa_scannet}, we can find that the base detector achieves close to 88\% performance of full supervision under one object per class. Even so, our method attains improvement and achieves 96\%  performance compared to the detector under full supervision.

\subsection{Expansion of Detectors} \label{sec:more_det}
To verify the extensibility of our method to other detectors, we employ additional detectors for indoor and outdoor scenarios. Specifically, FCAF3D~\cite{fcaf3d} and CenterPoint~\cite{centerpoint} are utilized for indoor and outdoor environments, respectively.
\begin{table*}
	\caption{Results on validation of ScanNet V2 and SUN RGB-D under sparse supervised setting.}
	\label{tab:sup_sca}
	\centering
	\resizebox{0.85\textwidth}{!}{
		\begin{tabular}{c|c|c|cc|cc}
			\toprule
			\multirow{3}{*}{Methods} & \multirow{3}{*}{Presented at} & \multirow{3}{*}{Paradigm} & \multicolumn{2}{c|}{ScanNet V2} & \multicolumn{2}{c}{SUN RGB-D}\\ 
			\cmidrule(r){4-7}
			& & & mAP@0.25 & mAP@0.5 & mAP@0.25 & mAP@0.5 \\
			\midrule
			FCAF3D~\cite{fcaf3d}&ECCV & Fully Supervised & 70.7 & 56.0  & 63.8 & 48.2 \\
			\midrule
			FCAF3D~\cite{fcaf3d}&ECCV & \multirow{2}{*}{Sparse Supervised} & 37.6 & 21.1  & 52.3 & 36.6 \\
			ours+FCAF3D &  - & &\textbf{54.6} & \textbf{36.6}  &\textbf{56.1} & \textbf{41.3}\\
			\bottomrule
	\end{tabular}}
\end{table*}

\begin{table*}
	\caption{Comparison with state-of-the-art methods on outdoor dataset under the sparse supervised setting.}
	\label{tab:sup_kitti}
	\centering
	\resizebox{0.85\textwidth}{!}{
		\begin{tabular}{c|c|c|ccc|ccc}
			\toprule
			\multirow{2}{*}{Methods}  & \multirow{2}{*}{Present at} & \multirow{2}{*}{Paradigm} & \multicolumn{3}{c|}{Car-3D AP (R40)} & \multicolumn{3}{c}{Car-BEV AP (R40)}\\ 
			\cmidrule(r){4-9}
			&	&  & Easy & Moderate &Hard& Easy & Moderate & Hard \\
			\midrule
			CenterPoint \cite{centerpoint} & CVPR & Fully Supervised&  89.0 & 80.5 &  76.5 & 92.9 & 89.0 & 87.5 \\ 
			\midrule
			CenterPoint~\cite{centerpoint}& CVPR & \multirow{2}{*}{Sparse Supervised} &  49.6 & 31.5  & 25.9 & 56.7 & 42.5 & 34.1 \\
			ours+CenterPoint & - &  &  \textbf{90.3}  &\textbf{76.3}  & \textbf{67.1}  & \textbf{94.9}  & \textbf{88.1}  & \textbf{78.9} \\
			\bottomrule
	\end{tabular}}
	\vspace {-5pt}
\end{table*}
To extend our method on indoor detector FCAF3D~\cite{fcaf3d}, we need two modifications. The first one is in the prototype-based object mining module, we add the centerness assisted with the origin propagation probability $W$ for the prototype labels generation. The other one is the multi-label cooperative refinement module, we add the centerness~\cite{fcaf3d} to select the pseudo labels. As shown in Tab.~\ref{tab:sup_sca}, our proposed method still significantly enhances performance when based on the FCAF3D detector. In terms of mAP@0.25, we achieve improvements of 17 and 3.8 compared to the base detector on ScanNet V2~\cite{scannet} and SUN RGB-D~\cite{sunrgbd}, respectively. In terms of mAP@0.5, our method shows improvements of 15.5 and 4.7 on ScanNet V2 and SUN RGB-D, respectively.

In the main paper, CenterPoint~\cite{centerpoint} is used as a one-stage detector of Voxel-RCNN~\cite{voxelrcnn}, so we can extend our method to it with no modifications. The results based on  CenterPoint are presented in Tab.~\ref{tab:sup_kitti}. The table compares the performance of CenterPoint under fully supervised and sparse supervised paradigms. While sparse supervised CenterPoint shows significantly reduced performance, our method (ours+CenterPoint) improves the results substantially, achieving metrics close to the fully supervised detector. These results for indoor and outdoor scenes demonstrate the scalability of our method.

\subsection{Prototype \textit{v.s.} GT Sampling} \label{sec:pro_gt}
In this section, we present a comparative evaluation of the GT-sampling-based approach and our proposed prototype-based label mining method. Our method achieves superior performance, with mAP@0.5 scores of 26.5 and 37.7 on the ScanNet V2 and SUN RGB-D datasets, respectively, significantly outperforming the GT-sampling-based approach, which yields mAP@0.5 scores of 21.2 and 32.1 on the same datasets. This performance gap can be attributed to the limitations of GT sampling, where instance pasting often leads to geometrically inconsistent contexts. In contrast, our prototype-based method addresses this issue by clustering features across all categories and performing feature matching between prototype features and scene instance features, thereby ensuring geometrically coherent contexts. 


\subsection{Indoor Per-class Evaluation} \label{sec:per_class}
Following the common experimental setups for indoor 3D object detection~\cite{tr3d, fcaf3d}, we report the results for each category on ScanNet V2~\cite{scannet} and SUN RGB-D~\cite{sunrgbd}. Tab. \ref{tab:25scan},  Tab. \ref{tab:50scan} report the results of 18 classes of ScanNet V2 at 0.25 and 0.5 IoU thresholds, respectively. Tab. \ref{tab:25sun},  Tab. \ref{tab:50sun} show 10 classes of SUN RGB-D results on 0.25 and 0.5 IoU thresholds, respectively. For Scannet V2 and SUN RGB-D, a comprehensive view of Tab. \ref{tab:25scan}, Tab. \ref{tab:50scan}, Tab. \ref{tab:25sun}, and Tab. \ref{tab:50sun}, shows that the improvement of our method is reflected in all categories. This improvement is achieved by using the components we proposed in this paper. 

\subsection{Quantitative Results} \label{sec:visual}
\begin{figure*}[!ht]
	\centering
	\includegraphics[width=1.0\textwidth]{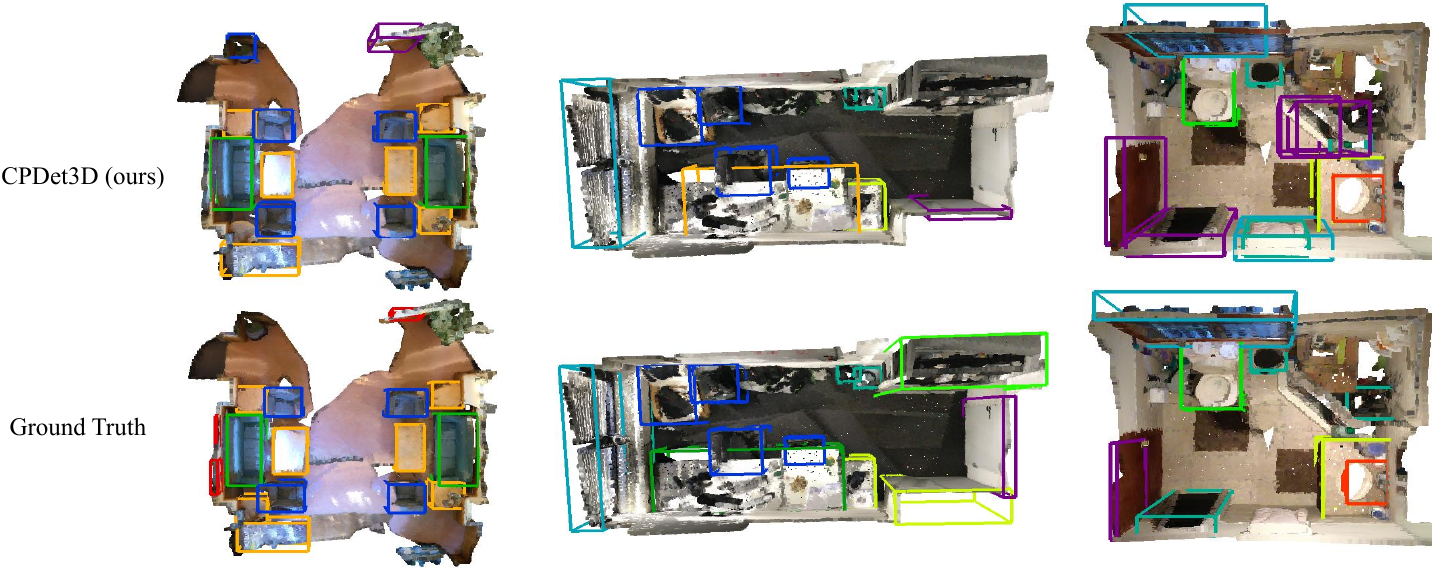}
	\caption{Visualization results of our method on the ScanNet V2 validation set trained under one object per scene setting.}
	\label{fig:scannet_show_sup}
\end{figure*}

\begin{figure*}[!ht]
	\centering
	\includegraphics[width=1.0\textwidth]{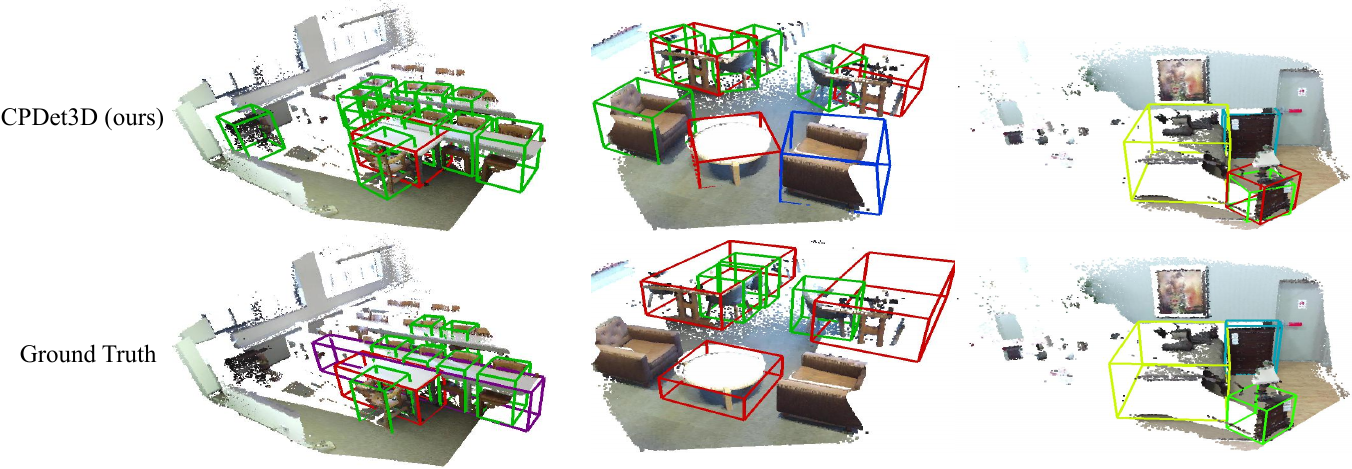}
	\caption{Visualization results of our method on the SUN RGB-D validation set trained under one object per scene setting.}
	\label{fig:sunrgbd_show}
\end{figure*}

\begin{figure*}[!ht]
	\centering
	\includegraphics[width=1.0\textwidth]{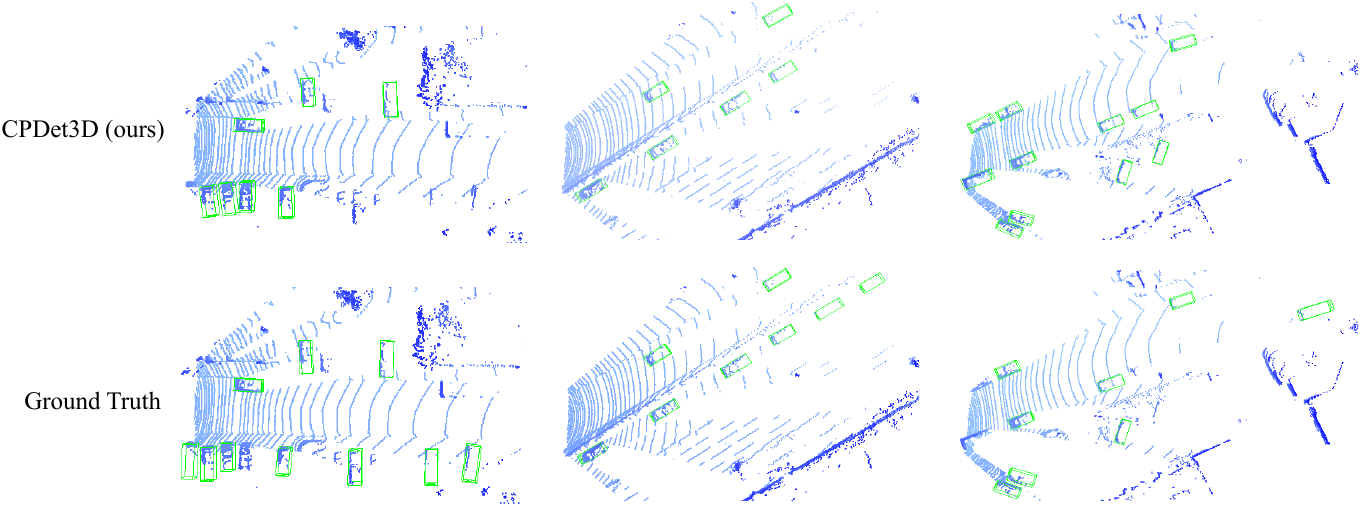}
	\caption{Visualization results of our method on the KITTI validation set trained under 10\% one object per scene setting.}
	\label{fig:kitti_show}
\end{figure*}

In this section, we provide more quantitative results on ScanNet V2~\cite{scannet}, SUN RGB-D~\cite{sunrgbd}, and KITTI~\cite{kitti}. The quantitative results of predicted bounding boxes on the ScanNet V2, SUN RGB-D and KITTI datasets are shown in Fig. \ref{fig:scannet_show_sup}, Fig. \ref{fig:sunrgbd_show}, and Fig. \ref{fig:kitti_show}, respectively. In particular, from Fig. \ref{fig:sunrgbd_show}, we can see that our proposed method can even detect some unlabeled objects in SUN RGB-D.

\section{Limitations} \label{sec:limitations}
Our method is designed to mine class-aware features from the entire dataset and assign prototype labels to unlabeled objects, followed by a simple iterative training process to progressively enhance model performance under sparse annotations. Experimental results demonstrate that our approach significantly improves the performance of the base detector across indoor and outdoor scenes. However, we have not yet explored network architecture design tailored for sparse-supervised 3D object detection, nor have we explicitly addressed scene-specific corner cases, such as hard instances at long distances. 

To tackle these challenges, future research could focus on redesigning the network to effectively mine features for challenging objects and developing adaptive data augmentation strategies to enhance the model’s performance in challenging cases. This field is still in its early stage of exploration and remains highly worthy of further research.

\begin{table*}[!th]
	\large
	\caption{3D detection scores per category on the ScanNet V2 under one object per scene sparse supervised setting, evaluated at 0.25 IoU threshold.}
	\centering
	\renewcommand{\arraystretch}{1.5}{
		\resizebox{1.0\textwidth}{!}{
			\begin{tabular}{c|c|c|c|c|c|c|c|c|c|c|c|c|c|c|c|c|c|c|c}
				\toprule
				Methods & cab & bed & chair & sofa & tabl & door & wind & bkshf & pic & cntr & desk & curt & frig & showr & toil & sink & bath & ofurn & mAP \\
				\midrule
				FCAF3D~\cite{fcaf3d} & 16.9 &48.2 & 80.9 & 51.4 &44.7 &25.9 &15.1 &27.6 &1.5 &17.3 &49.9 &19.6 &23.0 &38.1 &86.9 &42.0 &53.8 &34.5 &37.6 \\
				TR3D~\cite{tr3d}  & 11.1 &72.3 & 77.2 & 48.5 &34.6 &29.3 &11.1 &29.3 &4.5 &18.0 &48.6 &10.0 &28.1 &15.4 &83.2 &62.1 &62.1 &31.1 & 37.6 \\
				\midrule
				CPDet3D (ours) & \textbf{28.3} &\textbf{83.6} &\textbf{93.8} &\textbf{81.4} & \textbf{54.2} &\textbf{43.6} &\textbf{22.5} &\textbf{40.2} &\textbf{14.8} &\textbf{52.6} &\textbf{62.5} &\textbf{28.7} &\textbf{41.3} &\textbf{47.6} &\textbf{99.7} &\textbf{74.0} &\textbf{92.9} &\textbf{48.8} &\textbf{56.1}\\
				\bottomrule
	\end{tabular}}}
	\label{tab:25scan}
\end{table*}

\begin{table*}[h]
	\centering
	\caption{3D detection scores per category on the ScanNet V2 under one object per scene sparse supervised setting, evaluated at 0.50 IoU threshold.}
	\renewcommand{\arraystretch}{1.5}{
		\resizebox{1.0\textwidth}{!}{
			\begin{tabular}{c|c|c|c|c|c|c|c|c|c|c|c|c|c|c|c|c|c|c|c}
				\toprule
				Methods & cab & bed & chair & sofa & tabl & door & wind & bkshf & pic & cntr & desk & curt & frig & showr & toil & sink & bath & ofurn & mAP \\
				\midrule
				FCAF3D~\cite{fcaf3d} & 5.9 &31.1 & 67.0 & 35.5&31.2 &6.7 &1.1 &16.6 &0.0 &0.2 &26.4 &5.8 &16.8 &2.0 &79.9 &11.1 &19.0 &22.6 &21.1 \\
				TR3D~\cite{tr3d}  & 3.0 &52.4 & 64.1 & 33.3 &26.5 &13.6 &0.3 &15.3 &0.0 &2.0 &31.3 &0.4 &23.2 &0.4 &62.7 &16.8 &27.1 &19.5 &21.8 \\
				\midrule
				CPDet3D (ours) & \textbf{13.1} &\textbf{74.0} &\textbf{85.6} &\textbf{68.3} & \textbf{47.2} &\textbf{25.9} &\textbf{5.0} &\textbf{28.2} &\textbf{7.6} &\textbf{24.3} &\textbf{41.6} &\textbf{9.3} &\textbf{29.1} &\textbf{10.5} &\textbf{92.8} &\textbf{43.4} &\textbf{89.2} &\textbf{38.2} &\textbf{40.8}\\
				\bottomrule
	\end{tabular}}}
	\label{tab:50scan}
\end{table*}

\begin{table*}[!th]
	\large
	\caption{3D detection scores per category on the SUN RGB-D under one object per scene sparse supervised setting, evaluated at 0.25 IoU threshold.}
	\centering
	\renewcommand{\arraystretch}{1.5}{
		\resizebox{0.7\textwidth}{!}{
			\begin{tabular}{c|c|c|c|c|c|c|c|c|c|c|c}
				\toprule
				Methods &bath &bed &bkshf &chair &desk &dresser &nstand &sofa &table &toilet &mAP \\
				\midrule
				FCAF3D \cite{fcaf3d} &69.5 &81.7 &27.2 &63.5 &21.0 &24.6 &54.1 &59.0 &33.2 &88.8 &52.3 \\
				TR3D \cite{tr3d} &64.8 &83.8 &25.6 &67.1 &27.2 &25.6  &56.1 &58.6 &38.8 &91.6 &53.9 \\
				\midrule
				CPDet3D (ours) &\textbf{74.6} &\textbf{87.3} &\textbf{30.4} &\textbf{75.7} &\textbf{27.6} &\textbf{31.9} &\textbf{69.4} &\textbf{65.1} &\textbf{65.1} &\textbf{89.0} &\textbf{60.2} \\
				\bottomrule
	\end{tabular}}}
	
	\label{tab:25sun}
\end{table*}

\begin{table*}[!th]
	\large
	\caption{3D detection scores per category on the SUN RGB-D under one object per scene sparse supervised setting, evaluated at 0.50 IoU threshold.}
	\centering
	\renewcommand{\arraystretch}{1.5}{
		\resizebox{0.7\textwidth}{!}{
			\begin{tabular}{c|c|c|c|c|c|c|c|c|c|c|c}
				\toprule
				Methods &bath &bed &bkshf &chair &desk &dresser &nstand &sofa &table &toilet &mAP \\
				\midrule
				FCAF3D~\cite{fcaf3d} &52.4 &59.6 &8.8 &49.0 &6.8 &14.4 &42.4 &46.2 &19.0  &67.1 &36.6\\
				TR3D~\cite{tr3d} &43.0 &58.3 &7.8 &52.8 &8.7 &16.9 &41.8  &45.3 &21.1  &67.4 &36.3\\
				\midrule
				CPDet3D (ours) &\textbf{74.6} &\textbf{67.6} &\textbf{9.5} &\textbf{63.5} &\textbf{9.4} &\textbf{20.7} &\textbf{57.4} &\textbf{53.3} &\textbf{31.7} &\textbf{67.6} &\textbf{43.3} \\
				\bottomrule
	\end{tabular}}}
	\label{tab:50sun}
\end{table*}

\end{document}